 \pdfoutput=1

\documentclass[spconf]{article}
\usepackage{spconf,amsmath,epsfig,float,subfigure,multirow,graphicx,hyperref}
\usepackage{url}
 \usepackage{mathrsfs}
 \usepackage{amsfonts}
  \usepackage{array}
\begin{document}
\sloppy

\def\x{{\mathbf x}}
\def\L{{\cal L}}

\title{Interactive Deep Colorization With Simultaneous Global and Local Inputs}


\name{Yi Xiao$^1$, Peiyao Zhou$^1$, Yan Zheng$^2$}
\address{
1 College of Computer Science and Electronic Engineering\\
2 College of Electric and Information Engineering\\
Hunan University\\
Changsha, China\\
}



\maketitle

\begin{abstract}
 Colorization methods using deep neural networks have become a recent trend. However, most of them do not allow user inputs, or only allow limited user inputs (only global inputs or only local inputs), to control the output colorful images. The possible reason is that it's difficult to differentiate the influence of different kind of user inputs in network training. To solve this problem, we present a novel deep colorization method, which allows simultaneous global and local inputs to better control the output colorized images.  The key step is to design an appropriate loss function that can differentiate the influence of input data, global inputs and local inputs. With this design, our method accepts no inputs, or global inputs, or local inputs, or both global and local inputs, which is not supported in previous deep colorization methods. In addition, we propose a global color theme recommendation system to help users determine global inputs. Experimental results shows that our methods can better control the colorized images and generate state-of-art results.


     \end{abstract}


\section{Introduction}
Image colorization refers to the technique that adds colors to monochrome images or videos.
Generally speaking, colorization is a ill-posed problem, it does not have a unique solution.
To get satisfactory colorized results, two categories of method have been proposed: user-guided edit propagation and data-driven automatic colorization.

The user-guided edit propagation methods~\cite{Levin2004Colorization,HORIUCHI-Example-2005,Huang2005An,yatziv2006fast,Luan2007Natural,Kawulok-2010,Yao-Bayesian-2011,Kawulok-Textural-2011}  require the user to draw colored strokes
and propagate the colors across the image by solving a global optimization problem.
 These methods can achieve impressive colorized images, but often require a very large number of scribbles for images with complex textures. This is because each different color
region must be explicitly marked by a different colored stroke, even regions with obvious semantic hints, such as a blue sky or green trees, need to be specified by the user.

To address this problem, early data-driven colorization methods proposed to automatically colorize a grayscale image by learning the color hints from one or several exemplar color images with similar semantics~\cite{Welsh2002Transferring,Chang2005Example,Chia2011Semantic,Gupta2012Image,liu-2008-intrinsic}. Unfortunately, it may be time-consuming or hard to find a suitable exemplar image sometimes. With the popularity of deep learning, recent data-driven colorization methods using deep neural networks have become a recent trend~\cite{Cheng2015Deep,DBLP:journals/corr/NguyenMT16,Limmer2017Infrared}.
Using a large number of grayscale and color image pairs, the deep colorization methods learn a parametric mappings for fully automatic colorization. These methods can generate plausible colorful images in most of time. However, since an semantic region can have multiple choices of colors, the results can contain colors or styles which users do not expected. For example, users may want a green mountain in spring, but get a yellow mountain in autumn.

The excellent recent work, user-guide deep colorization,  by Zhang et al~\cite{Efros2017Real} combines the advantages of user-guided and data-driven methods. It provide better user controls by taking either global inputs or local inputs in the deep network training. The color of the colorized image can be controlled by a global color histogram, or a few local color points. The user-guide deep colorization~\cite{Efros2017Real} can generate plausible colorized images according to the users expectation with only a few inputs. However, The user-guide deep colorization method does not support simultaneous global inputs and local inputs, it can only allow one kind of inputs at one time.

We argue that supporting simultaneous global inputs and local inputs can provide better control on the output images. The ideal case is that a user can control the overall color style of the image with global color inputs, and meanwhile assign local colors to certain regions with local inputs.

Supporting multiple kinds of inputs simultaneously in deep networks is not straightforward. It's difficult to differentiate the influence of different kind of user inputs in network training. An example is shown in Figure~\ref{fig:LocalLoss}, the influence of local inputs is lost if the loss function is not designed appropriately.  To solve this problem, we present a novel deep colorization method, which allows simultaneous global and local inputs to better control the output colorized images. Our basic model neural network is a U-net network including three parts, the feature extraction part, the fusion part, and the reconstruction part. The feature extraction part, consisting of 3 convolution layers and 3 pooling layers, extracts features from the gray-scale images and the local input images. The fusion part fuse the global inputs (color themes) and the extracted feature maps. The reconstruction part, consisting 3 deconvolution layers, then reconstructs the two chrominance channels ({\em Lab} space). The key step is to design an appropriate loss function that can differentiate the influence of input data, global inputs and local inputs. With this design, our method accepts no inputs, or global inputs, or local inputs, or both global and local inputs, which is not supported in previous deep colorization methods.

Different with previous works, which use exemplar images or histograms as global inputs, we choose to use the color theme, a template of colors possibly associated verbal description~\cite{Wang:2010:DIC:1882262.1866172}, as the global input. It is more easier for users to assign a color theme than to choose exemplar images or histograms. In our model, we accept color themes consist of 3 - 7 colors. More colors are possible but difficult to use. To further save work time for users, we propose a global color theme recommendation system to help users determine the global input. Inspired by~\cite{Wang:2010:DIC:1882262.1866172}, we build a mapping between grayscale texture features and color histograms from a large image data base, and use the mapping to predict the color themes. Also, the user can define its own color theme.

\begin{figure*}[t]
\centering
	\begin{tabular} {@{\extracolsep{1mm}}c @{\extracolsep{1mm}}c @{\extracolsep{1mm}}c @{\extracolsep{1mm}}c @{\extracolsep{1mm}}c @{\extracolsep{1mm}}c @{\extracolsep{1mm}}c @{\extracolsep{1mm}}c}
		{ } &
		{ } &
		{ } &
		\includegraphics[height = 0.36cm]{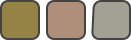} &
		\includegraphics[height = 0.36cm]{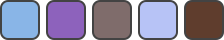} &
		\includegraphics[height = 0.36cm]{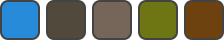} &
		{} &
		\includegraphics[height = 0.36cm]{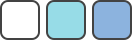} \\
		\includegraphics[width=0.12\textwidth]{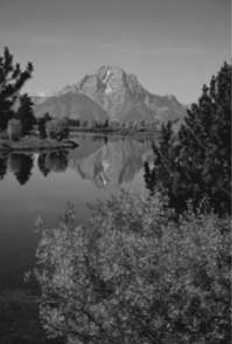} &
		\includegraphics[width=0.12\textwidth]{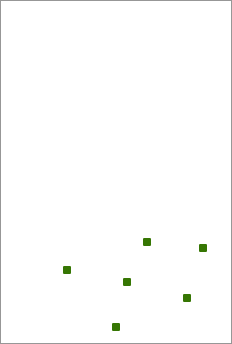} &
		\includegraphics[width=0.12\textwidth]{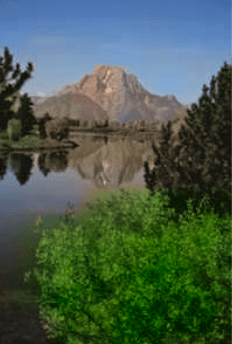} &
		\includegraphics[width=0.12\textwidth]{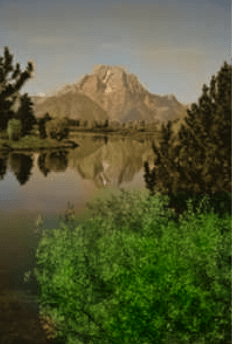} &
		\includegraphics[width=0.12\textwidth]{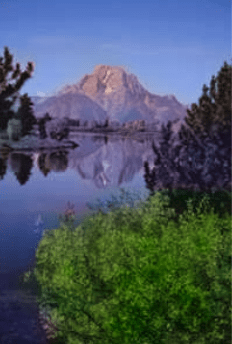} &
		\includegraphics[width=0.12\textwidth]{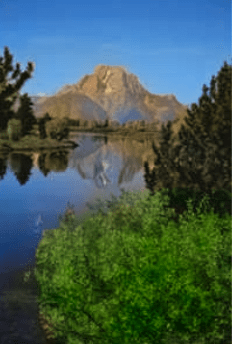} &
		\includegraphics[width=0.12\textwidth]{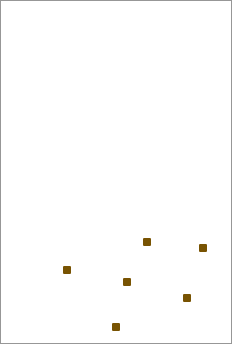} &
		\includegraphics[width=0.12\textwidth]{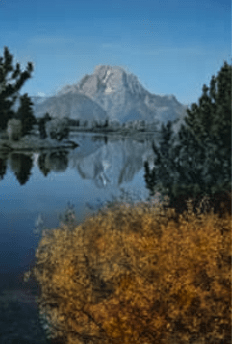} \\
		{ } &
		{ } &
		{ } &
		\includegraphics[width=0.12\textwidth]{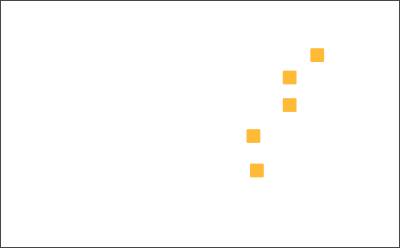} &
		\includegraphics[width=0.12\textwidth]{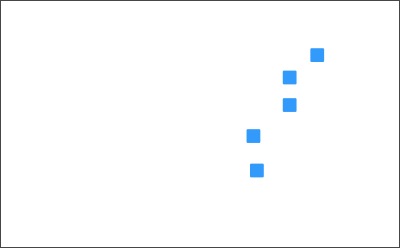} &
		\includegraphics[width=0.12\textwidth]{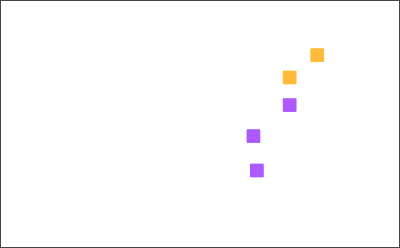} &
		\includegraphics[width=0.12\textwidth]{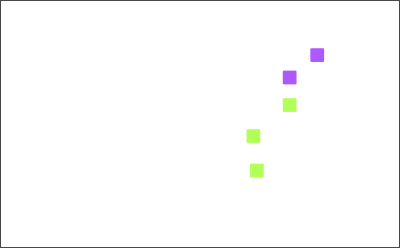} &
		\includegraphics[width=0.12\textwidth]{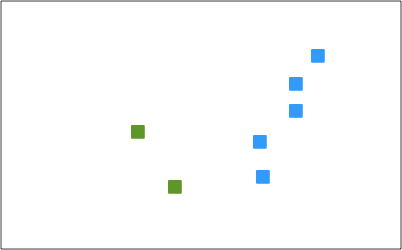} \\
		\includegraphics[width=0.12\textwidth]{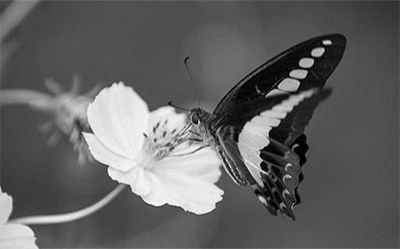} &
		\includegraphics[height = 0.36cm]{figures/Display/312.png} &
		\includegraphics[width=0.12\textwidth]{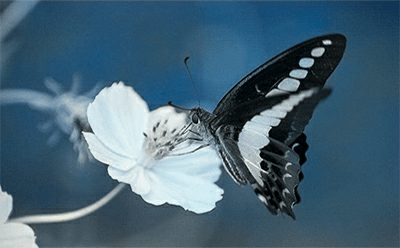} &
		\includegraphics[width=0.12\textwidth]{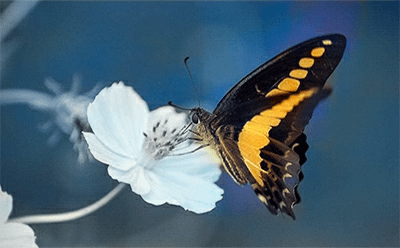} &
		\includegraphics[width=0.12\textwidth]{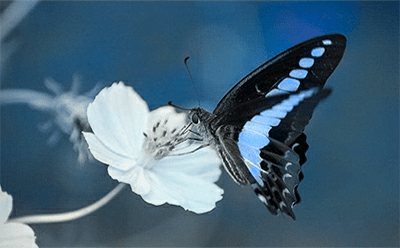} &
		\includegraphics[width=0.12\textwidth]{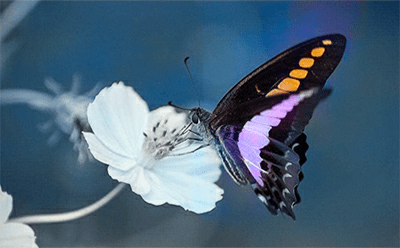} &
		\includegraphics[width=0.12\textwidth]{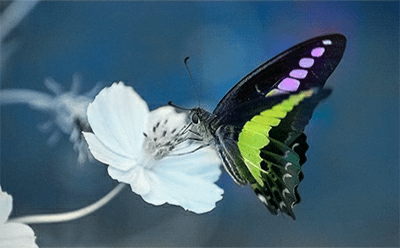} &
		\includegraphics[width=0.12\textwidth]{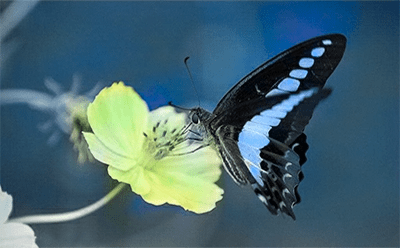}\\
	\end{tabular}
	\caption{ Our method colorizes two gray images by global input and local input simultaneously. The third column are the results using only local or global inputs. The other results are obtained using  simultaneously global input and local input.  }\label{fig:teaser}
\end{figure*}

Experimental results show that the color images generated by our method look real and natural, and the detail of the image are preserved well. Using different color themes and local inputs, we can output different style images (See Figure~\ref{fig:teaser}). To summarize, our methods can better control the colorized images and generate state-of-art results.

Our contributions in this paper are as follows:
\begin{itemize}
\item We propose a novel deep colorization methods that supports no inputs, or global inputs, or local inputs, or both global and local inputs.
\item Our methods enables global inputs using color themes of variable color numbers.
\item We propose a color theme recommendation system which can suggest the color themes for users.
\end{itemize}

\section{Related Work}

Image colorization is always a hot topic in the filed of image process. After the advent of deep convolution neural network, it is devoted to train images on network to achieve the goal of great colorization. At present, there are two main ways of colorization. One is user-guided edit propagation colorization, and the other is automatic colorization.

About user-guided coloring method, most of them are devoted to add scribbles on gray images. In the early paper, like \cite{Levin2004Colorization}, it is necessary for users to draw some desired color graffitis on certain regions, through the color propagating among pixels which are similar in intension to achieve the effect of colorization. \cite{Huang2005An} and \cite{Li2008ScribbleBoost} worked on the issue of edge detection and enhancement, which can improve the quality of color transmission. In ~\cite{Luan2007Natural}, users select the area of interest and then specify some color points within the area. ~\cite{Xu09sigasia} and ~\cite{Chen2012Manifold} are optimized for edit propagation and shading. It's also popular to use color theme, such as using data driven color theme enhancements in ~\cite{Wang:2010:DIC:1882262.1866172}, which is used to change the color style of color images through different color themes. We take this method and using color theme as global input to color gray images.  ~\cite{Chang:2015:PPR} is a popular method to use the palette to change the corresponding color on color image as same as ~\cite{Wang:2010:DIC:1882262.1866172}. In \cite{Endo2016DeepProp}, it is proposed to automatically learn the degree of similarity between user strokes and input images to propagate graffiti color. \cite{sangkloy2016scribbler} uses generative adversarial networks to achieve transforming sketch to real images, which also supports to add user strokes. In \cite{Efros2017Real}, by adding global features and local features on deep convolution neural network to train a colorization model, separately, users can specify the color at any location to render the color, or use a reference image to transfer global color. Our method can achieve colorization using both of them simultaneously. There are online applications developed mainly for line-drawing colorization, like ~\cite{DBLP:journals/corr/Frans17} and ~\cite{paintschainer.preferred.tech}.

For more convenient to color grayscale images globally, there are some method using reference images as global inputs early. \cite{Erik2001Color} transfers colors between images with a simple algorithm. \cite{Welsh2002Transferring} and \cite{Chang2005Example} transmit the color emotion of reference images to grayscale images by matching the brightness and texture information between images. \cite{Chia2011Semantic} needs users to provide the semantic information of the foreground image. The system searches for and downloads pictures with the same foreground semantics on the network, and use them to color the foreground and the background. \cite{Gupta2012Image} can quickly transfer the color of a reference image to the target image by using a fast cascade feature matching scheme to exploit multiple image features.

But using reference images which is similar with target images to render the color needs a large image database. Users also have to spend a lot of time to choose satisfactory reference image. So, the research of automatic colorization gets more and more attention. \cite{Li2007Learning} uses algorithm to reserve grayscale images or videos. In this method, some representative color points are reserved and used to restore these medias. \cite{Charpiat2008Automatic} uses machine learning tools to extract as much information as possible from the color sample dataset and then estimates the probability color distribution for every pixel. \cite{Morimoto2009Automatic} matches grayscale image with reference images downloaded automatically from web to transfer color. \cite{Deshpande2015Learning} train the objective function based on image features on the LEARCH framework and achieve the coloring effect by minimizing the objective function. \cite{Cheng2015Deep} introduces the concept of feature descriptor, and take the feature descriptors extracted from the grayscale image as input, and finally output the corresponding color values of the UV channel. In \cite{DBLP:journals/corr/NguyenMT16}, a convolutional neural network (CNN) which accepts black and white images as input is designed and constructed, and a statistical learning driven method is used to solve the problem of grayscale colorization. The end-to-end network in \cite{IizukaSIGGRAPH2016} adds global feature and uses classification label to optimize output results. In addition, \cite{Zhang2016Colorful}, \cite{Varga2017Fully}, \cite{Limmer2017Infrared}, ~\cite{Zhao2016Retracted}, \cite{larsson2016learning} also achieve the effect in coloring gray image without users' intervention. \cite{DBLP:journals/corr/IsolaZZE16} uses conditional adversarial networks and \cite{DBLP:journals/corr/Frans17} proposes a setup utilizing two networks in tandem to achieve colorization, which both are worth thinking about.

Neural network also has shown us some surprising results in other fields of image processing. ~\cite{Liu2014AutoStyle} reaches the target of automatic style transferring between images. As well as large-scale image recognition ~\cite{Simonyan2014Very},  automatic photo adjustment ~\cite{Yan2014Automatic}, sketch simplification ~\cite{Simo2016Learning}, context encoders ~\cite{pathakCVPR16context} and cartoon colorization ~\cite{Varga2017Automatic}. Accuracy and practicability of neural network is the reason that we chose neural network to research and construct satisfactory colorization method.

\section{Our Method}

\begin{figure*}
  \centering
  \includegraphics[width=0.9\textwidth]{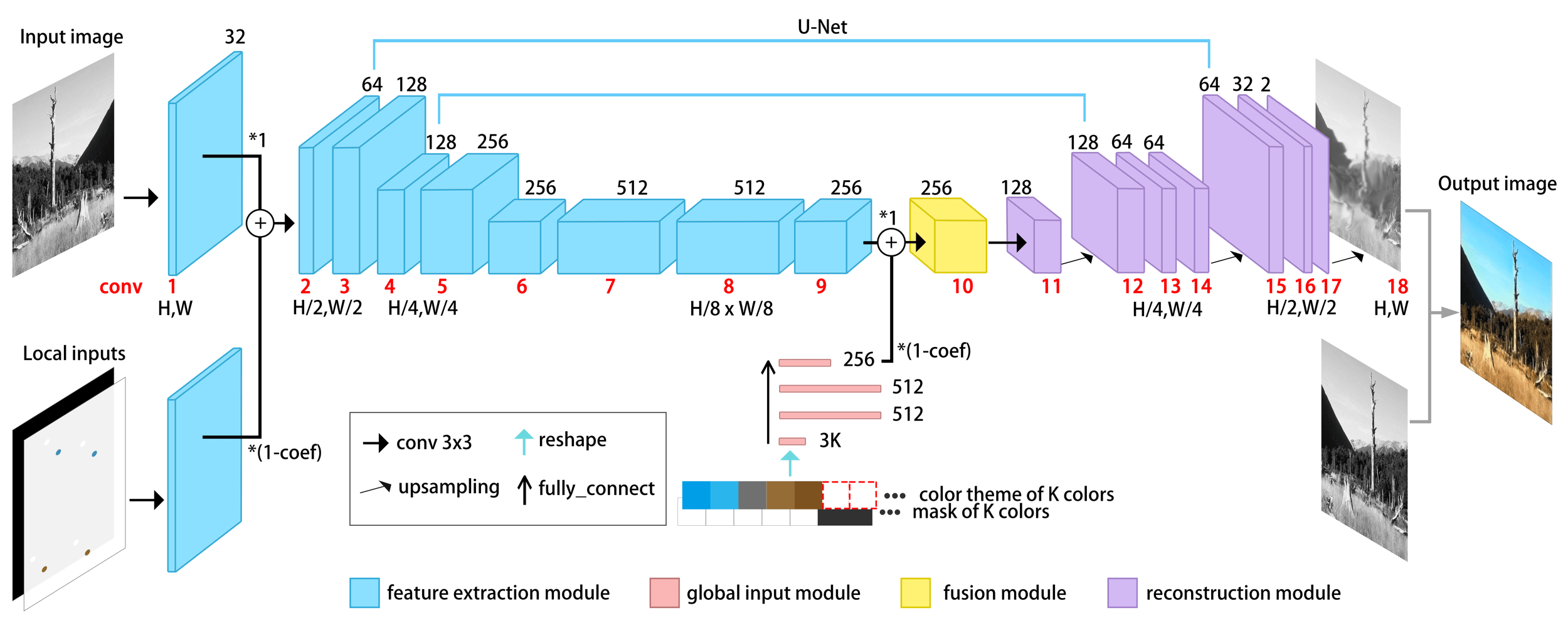}\\
  \caption{Our network model for colorization. The steps between layers without arrow instruction are convolution with $3\times3$ kernel.}\label{NetworkModel}
\end{figure*}

Our deep network model takes in a gray-scale image ({\em L} channel in CIE {\em Lab} space), a global input and a local input and output the corresponding {\em ab} channels. The structure of the model is shown in Figure~\ref{NetworkModel}. The global input includes a color theme of several colors and a mask indicating the number of colors (all black means no global inputs). The local input includes a image consisting of the assigned local color points and a mask indicating the position of color points (all black means no local inputs). Therefore, by setting the masks of global inputs and local inputs, our model accepts four combination of user inputs: no inputs, global inputs, local inputs, global inputs and local inputs.




\subsection{Problem Formulation}
The inputs of our model are a grayscale image $X\in { R }^{ H\times W\times 1 }$,
 a global input tensor $U_g\in R^{1 \times K \times 3}$, and a local input tensor $U_l \in R^{H\times W\times 3}$, where $H$, $W$ is the width and height of the input image, and $K$ is the number of colors in the color theme. The output is a tensor $O \in R^{ H\times W\times 2 }$.
We aims to train a convolution neural network (CNN), denoted by $\mathcal{F}(X,U_g,U_l;\theta)$, to approximate the mapping between gray and color, under the constraint of user inputs. Therefore, the colorization problem can be formulated as

\begin{equation}
\theta^* = \mbox{arg} \min_{\theta} \mathbb{E}_{X,U_g,U_l,Y,\mathscr{D}}[ \mathcal{L}(\mathcal{F}(X,U_g,U_l;\theta),U_g, U_l, Y)],
\end{equation}

where $\mathscr{D}$ denotes the training data set, $\mathcal{L}$ denotes the loss function, and $Y \in R^{ H\times W\times 2 }$ is ground truth image color. More details of the loss function will be presented in Section~\ref{sec:LossFun}.
It should be noticed that we train a single model to handle simultaneous global input and local input. Moreover, our loss function is explicitly related to the global input and the local input. These are the two key points which differ our method from \cite{Efros2017Real}.

\subsection{User inputs}
\label{Sec:inputs}
\noindent\textbf{Global inputs:}
To control the overall color style of the output, we design the global input as a color theme with $K$ colors plus a one-channel mask indicating the number of colors. An example of color theme with 5 colors and the corresponding mask are shown in Figure~\ref{NetworkModel}. To prepare training data for the global input, we use the K-mean clustering algorithm to find the $K$ representative colors for each color image in the data set. The $K$ representative colors form the global color theme. To enable color theme with variable number of colors, $K$ randomly varies in the interval [3,7]. Color theme with more colors are also supported by modifying the interval, but may be harder to use. The {\em ab} channels of each color theme $U_g^c\in R^{1 \times K \times 2}$ and its mask  $M_g\in R^{1 \times K \times 1}$ forms the global input $U_g = \{U_g^c, M_g\}\in R^{1 \times K \times 3}$

\begin{figure}
  \centering
  \includegraphics[width=0.45\textwidth]{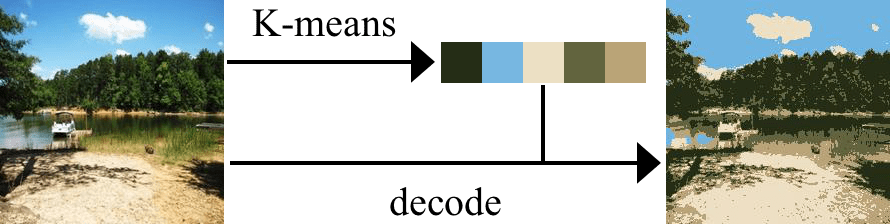}\\
  \caption{Generation process of 5-color-map image}\label{KColorMap}
\end{figure}

For each color image, we generate a $K$-color map by decoding the color image with its representative colors. Figure~\ref{KColorMap} shows an example. The {\em ab} channels of $K$-color map, $I\in R^{H \times W \times 2}$ is used to calculate the loss function which measures the similarity between the color theme and the output image. More details of loss functions will be described in Section~\ref{sec:GlobalLossFun}.



\noindent\textbf{Local inputs:}\label{sec:LocalInput}
Similar to \cite{Efros2017Real}, we prepare the training data for the local inputs colors $U_l^c\in { R }^{ H\times W\times 2 }$ by randomly selecting some points from the {\em ab} channels of every color image. If there are no samples in certain position, the {\em ab} values are set to zeros. A one-channel mask $M_l \in { R }^{ H\times W\times 1 }$ is also generated to indicate the positions of the local inputs.
"1" in certain position means there is an input there, while "0" means there are no inputs in the position. Finally, the local inputs $U_l = \{U_l^c, M_l\} \in { R }^{ H\times W\times 3}$.

\subsection{NetWork model}
As shown in Figure~\ref{NetworkModel}, our network model uses a U-Net structure~\cite{Ronneberger2015U}, which is already shown to works well in colorization~\cite{DBLP:journals/corr/IsolaZZE16,Efros2017Real}.
It is mainly composed of four parts: the feature extraction module, the global input module, the fusion module and the reconstruction module.


\subsubsection{Feature extraction module}
The feature extraction module corresponds to the blue part in Figure~\ref{NetworkModel}.
The feature extraction module takes in a grayscale image $X\in { R }^{ H\times W\times 1 }$ and a local input $U_l\in { R }^{ H\times W\times 3}$. Initially, $X$ and $U_l$ are respectively convolved to a tensor of size $H\times W\times 32$. These two tensors are merged to a single tensor of size $H\times W\times 32$ by linear interpolation, whose weight is treated as a parameter of the network to be trained. The merged tensor is then feeded to the following convolution blocks.

In convolution block 2 to 7, all the convolution kernels are  $3\times 3$. Through every layer, only the tensor sizes of input are halved spatially with the stride $2\times 2$, or only the tensor dimensions are doubled with the stride $1\times 1$. After conv7, the tensor is $\frac { H }{ 8 } \times \frac { W }{ 8 } \times 512$. We use convolution layers with stride $2\times 2$ instead of using max-pooling layers to reduce the tensor sizes, which is helpful in increasing the spatial support of each layer~\cite{IizukaSIGGRAPH2016}.  Conv8 and conv9 process the tensor further with convolution layers to reduce the 512-channel tensors to 256-channel tensor. The output of feature extraction module is a tensor of size $\frac { H }{ 8 } \times \frac { W }{ 8 } \times 256$, which will be fused with the global input.

\subsubsection{Global input module}
The global input module, the pink part in Figure~\ref{NetworkModel}, is one of the characteristics of our method. It takes in the global inputs $U_g \in R ^{ 1\times K \times 3 }$, which consists of the {\em ab} channels of the color theme $U_g^c$ and the corresponding mask $M_g$. To to unify its size with the size of the fusion module, we reshape it to a tensor of size $1\times1\times 3K$, which is then processed by three fully connected layers. This operation is similar to the global feature network in~\cite{IizukaSIGGRAPH2016}, but we use it to handle user global inputs instead of the features exacted from the input gray-scale image.


\subsubsection{Fusion module}
The fusion layer, shown with yellow color in Figure~\ref{NetworkModel}, is an important step to fuse the global input and the extracted features. Similar operations are also used in~\cite{IizukaSIGGRAPH2016,Efros2017Real}. The global input is fused with the output of the feature extraction module by linear interpolation, whose weight is also treated as a parameter in the network to be trained. Finally the fusion module outputs a feature tensor of size $\frac { H }{ 8 } \times \frac { W }{ 8 } \times 256$ .

\subsubsection{Reconstruction module}

After the fusion module, the feature tensor is processed by a set of convolution layers and upsampling layers for {\em ab} channel reconstruction, shown with purple color in Figure~\ref{NetworkModel}. Convolution layers reduce tensor dimensions by half and upsampling layers doubles the wide and height of the tensor. Conv17 is the last convolution layer with a Sigmoid transfer function, followed by an upsampling layer. The advantage of Sigmoid is that the output range is limited. The output tensor is of size $H\times W \times 2$, which combines the input gray-scale image
 $X\in { R }^{H\times W\times 1}$ to generate the final output $O \in { R }^{ H\times W \times 3 }$.

\subsection{Loss function}
\label{sec:LossFun}
Designing an appropriate loss function is the most important part of our work. It is not straightforward to
design a loss function which can differentiate the impact of different inputs. We interpret the four combinations of inputs as follows:
\begin{itemize}
\item[1] No inputs means the user wants the colors to be assigned with the ``experiences" of training data.
\item[2] Only the global input means the user wants the colors to be conditionally assigned with the "experience" of training data.
\item[3] Only the local input means the user wants to assign his/her preferred color to certain region, but let the colors of other regions be assigned with the ``experiences" of training data.
\item[4] Both inputs mean that the user wants to assign his/her preferred color to certain region, but let the colors of other regions be conditionally assigned with the "experiences" of training data.
\end{itemize}




\subsubsection{Loss function for no input or global input}
\label{sec:GlobalLossFun}
\begin{figure}[h]
\centering
	\begin{tabular} {@{\extracolsep{1mm}}c @{\extracolsep{1mm}}c @{\extracolsep{1mm}}c @{\extracolsep{1mm}}c }
		{$\delta$ = 0.5} &
		\includegraphics[width=0.12\textwidth]{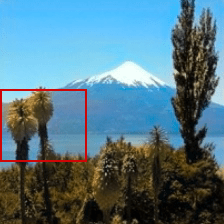} &
		\includegraphics[width=0.12\textwidth]{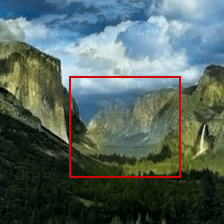} &
		\includegraphics[width=0.12\textwidth]{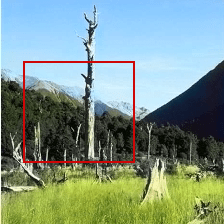} \\		
		{$\delta$ = 1.0} &
		\includegraphics[width=0.12\textwidth]{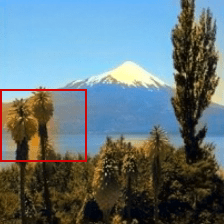} &
		\includegraphics[width=0.12\textwidth]{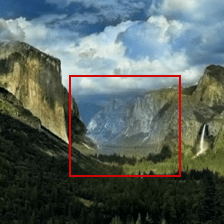} &
		\includegraphics[width=0.12\textwidth]{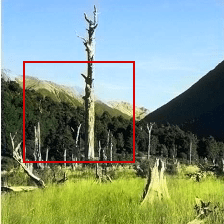} \\
	\end{tabular}
	\caption{Comparison of different delta parameters. The parts pointed out by red rectangle indicate the differences when use different delta parameters. }\label{delta parameters}
\end{figure}

For no inputs or only the global input, the straightforward choice is to measure the differences between {\em ab} channels of the output image $O\in { R }^{ H\times W \times 2 }$ and the ground truth color $Y\in { R }^{ H\times W \times 2 }$. There are many kind of loss functions as discussed in~\cite{Efros2017Real}. We choose the Huber loss as it produces relative high saturation effect. The Huber loss is given by
\begin{equation}\label{eq:HubberLoss}
\mathcal{L}_H(O,Y) = \left\{ {\begin{array}{*{20}{c}}
{\frac{1}{2}{{(O - Y)}^2}~\mbox{for}~\left| {O - Y} \right| \le \delta }\\ \\
{\delta \left| {O - Y} \right| - \frac{1}{2}{\delta ^2}   ~\mbox{otherwise}},
\end{array}} \right.
\end{equation}
where $\delta$ is the parameter of the Huber loss. The value of $\delta$ will slightly affect the results as shown in Figure~\ref{delta parameters}. We set $\delta = 0.5$, as this value works well in our experiments.


However, in some cases, the impact of the global input (color theme ) is not so obvious. Therefore, we add an additional part which calculates the Huber loss of the output image and the K-color map $I$, which is defined by decoding the ground truth with the color theme (see Section~\ref{Sec:inputs} and Figure~\ref{KColorMap}).

\begin{figure*}
\centering
	\begin{tabular} {@{\extracolsep{1mm}}c @{\extracolsep{5mm}}c @{\extracolsep{5mm}}c @{\extracolsep{1mm}}c @{\extracolsep{1mm}}c @{\extracolsep{1mm}}c @{\extracolsep{1mm}}c}
		\includegraphics[width=0.12\textwidth]{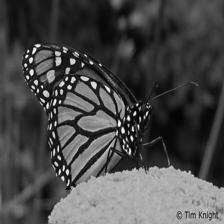} &
		\includegraphics[height=1.2cm]{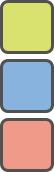} &
		\includegraphics[width=0.12\textwidth]{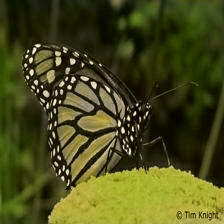} &
		\includegraphics[width=0.12\textwidth]{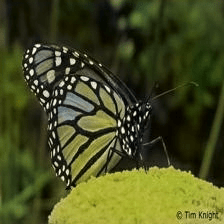} &
		\includegraphics[width=0.12\textwidth]{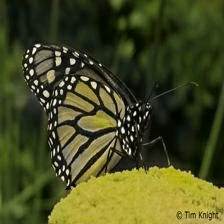} &
		\includegraphics[width=0.12\textwidth]{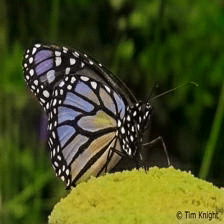} &
		\includegraphics[width=0.12\textwidth]{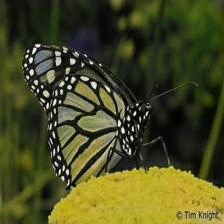}
		 \\
		
		\includegraphics[width=0.12\textwidth]{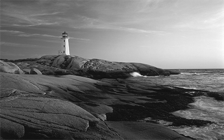} &
		{ } &
		\includegraphics[width=0.12\textwidth]{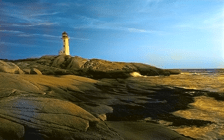} &
		\includegraphics[width=0.12\textwidth]{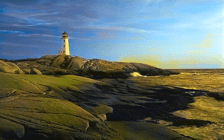} &
		\includegraphics[width=0.12\textwidth]{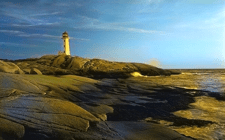} &
		\includegraphics[width=0.12\textwidth]{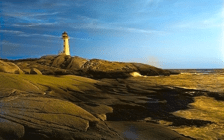} &
		\includegraphics[width=0.12\textwidth]{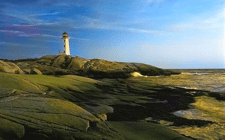}
		 \\		
		{ } &
		{ } &
		{(a) 0.1:0.9} & {(b) 0.3:0.7} & {(c) 0.5:0.5} & {(d) 0.7:0.3} & {(e) 0.9:0.1}

	\end{tabular}
	\caption{ Comparison of different fusion times and comparison of different ratios of loss. (a) to (e) show the results with different loss ratios.  (d) is the ratio we finally chosen for our method, which meets our demands for bright and well-distributed colorization and avoids color overflow.}\label{LossRatio}
\end{figure*}

Therefore, the loss function for this part is given by
\begin{equation}\label{eq:globalLoss}
{\mathcal{L}_g}  = \alpha_1*{\mathcal{L}_H(O,Y)} + \alpha_2*{\mathcal{L}_H(O,I)},
\end{equation}
where $\alpha_1$ and $\alpha_2$ are two parameters to balance the influence of two parts. The influence of the two parameters is shown in Figure~\ref{LossRatio}. As we expect that output images not only keep natural color but also reflect colors of the color theme, we set $\alpha_1=0.7$ and $\alpha_2=0.3$, with which the output can preserve the characteristics of the color theme with almost no color overflows, as showed in Figure~\ref{LossRatio}.
When there is no color theme input, the K-color map image will be replaced by the ground truth. In this case, ${\mathcal{L}}_g = {\mathcal{L}}_y$.

\subsubsection{Loss function for local input}
The local input in our method is $U_l = \{U_l^c, M_l\} \in { R }^{ H\times W\times 3}$, which is combined by the {\em ab} channels of the user input and the corresponding mask, as introduced in Section~\ref{sec:LocalInput}. If a user provides a local input in a certain position, we think the user wants the preferred colors instead of colors assigned by the ``experience" of training data. In there is only the local input, the Hubber loss $\mathcal{L}_H(O,Y)$ defined in Equation~(\ref{eq:HubberLoss}) is already shown to works well~\cite{Efros2017Real}, since the differences of the output image and the local inputs at positions with local inputs are already contained in Equation~(\ref{eq:globalLoss}).

\begin{figure}[h]
\centering
	\begin{tabular} {@{\extracolsep{1.5mm}}c @{\extracolsep{1.5mm}}c @{\extracolsep{1.5mm}}c @{\extracolsep{1.5mm}}c }	
        \includegraphics[width=0.11\textwidth]{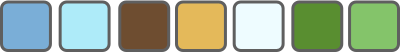} &
		{} &
		{} &
		{} \\
		\includegraphics[width=0.11\textwidth]{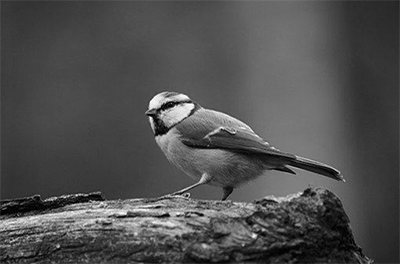} &
		\includegraphics[width=0.11\textwidth]{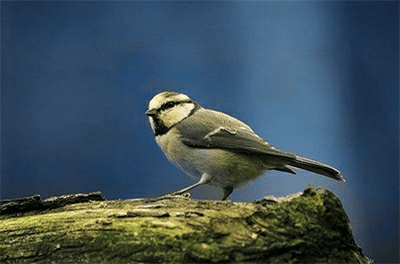} &
		\includegraphics[width=0.11\textwidth]{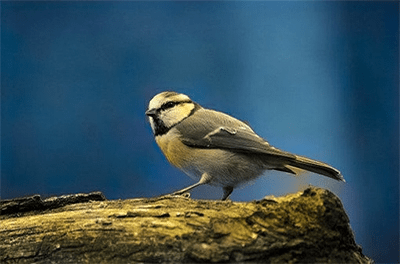} &
		\includegraphics[width=0.11\textwidth]{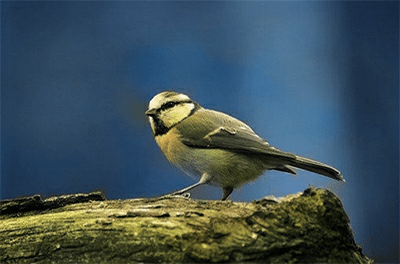} \\		
(a) Inputs & (b) only $\mathcal{L}_g$  & (c) ${\mathcal{L} _g}$+${\mathcal{L} _s}$ & (d) ${\mathcal{L} _g}$+${\mathcal{L} _s}$+${\mathcal{L} _p}$ \\	
\includegraphics[width=0.11\textwidth]{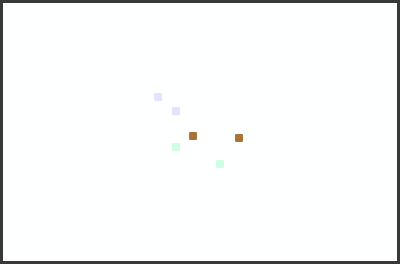} &
		\includegraphics[width=0.11\textwidth]{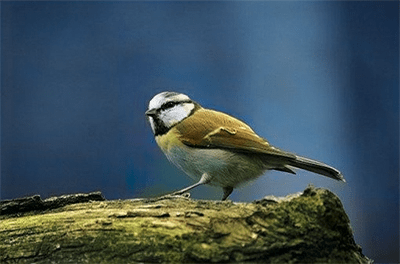} &
		\includegraphics[width=0.11\textwidth]{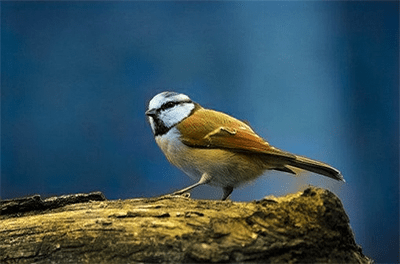} &
		\includegraphics[width=0.11\textwidth]{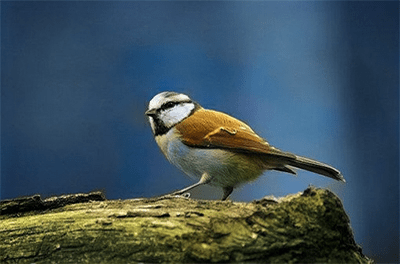} \\	
         (e) Inputs & (f) only $\mathcal{L}_g$ &(g) ${\mathcal{L} _g}$+${\mathcal{L} _s}$ & (h) ${\mathcal{L} _g}$+${\mathcal{L} _s}$+${\mathcal{L} _p}$ \\
		\includegraphics[width=0.11\textwidth]{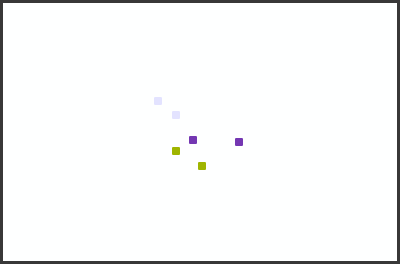} &
		\includegraphics[width=0.11\textwidth]{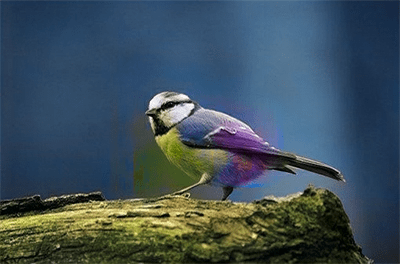} &
		\includegraphics[width=0.11\textwidth]{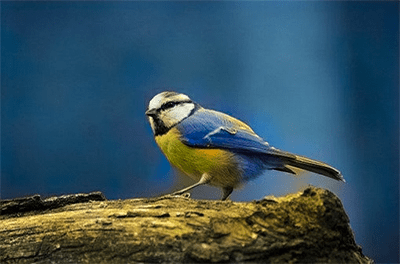} &
		\includegraphics[width=0.11\textwidth]{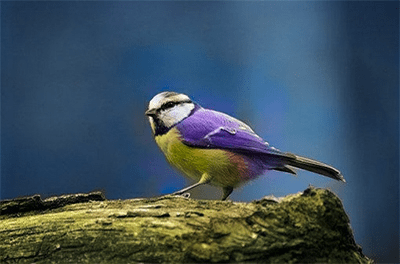} \\
		(i) Inputs & (j) only $\mathcal{L}_g$ & (k) ${\mathcal{L} _g}$+${\mathcal{L} _s}$ & {(l) ${\mathcal{L} _g}$+${\mathcal{L} _s}$+${\mathcal{L} _p}$} \\
	\end{tabular}
	\caption{Comparison of different combination of loss functions.  (a) is a grayscale image. We use the color theme above (a) as global input. (b) to (d) are results of using the color theme only. The last two rows show results of using both the color theme and local inputs, which are different in color of some points. Subscripts of images show the composition of loss function used. For example, (b) is the result of only using $\mathcal{L}_g$ as loss function. (c) add $\mathcal{L} _s$. (d) add $\mathcal{L} _s$ and $\mathcal{L} _p$, which is our final selection loss function. }\label{fig:LocalLoss}
\end{figure}

However, if there are simultaneous global input and local input, the case becomes more complex.
The Hubber loss may seems to work sometimes, when the local inputs are consistent or close with the global color theme, as shown in Figure~\ref{fig:LocalLoss}(f). For example, the input brown color in the back and swing is close to one color in the color theme (Figure~\ref{fig:LocalLoss}(a)). In this case, the local input can enhance the colors of local regions compared to results with only global inputs (Figure~\ref{fig:LocalLoss}(b)). But there exist some colors which overflow from the bird body to the background (TODO, indicate in the figure).

The color overflow can be removed by adding the gradient of the output image and the ground truth in the loss function, given by
\begin{equation}\label{eq:SobelLoss}
\mathcal{L}_{ s }=\mbox{MSE}  \left( { O }_{ Sobel }, { Y }_{ Sobel } \right),
\end{equation}
 where MSE denotes the mean squared error, Sobel denotes the Sobel gradient operator. As shown in Figure~\ref{fig:LocalLoss}(g), the color overflows are removed at the cost of removing the local color impacts on the head and abdomen.

When the local input colors are quite different with the colors in the global color theme, the phenomenon is more obvious, as shown in Figure~\ref{fig:LocalLoss}(j)(k). Using only the loss $L_g$ cannot guarantee the local color impact, and results in more obvious color overflows (Figure~\ref{fig:LocalLoss}(j)). The gradient loss $\mathcal{L}_{ s }$ almost remove the color overflows, but also remove the impact of local color inputs (Figure~\ref{fig:LocalLoss}(k)).
Since we expect the colorized image preserves not only the global input color theme, but also the local input colors, without few color overflows. We emphasize the impact the local inputs by adding the loss of the output image and the local inputs at positions with local inputs, given by
\begin{equation}\label{eq:LossLocalPoints}
\mathcal{L}_{ p }=MSE\left( \left( O\ast { M }_{ l } \right), \left( { U }_{ l }^{ c }\ast { M }_{ l } \right)  \right)
\end{equation}
where MSE denotes the mean squared error, $M_l$ is the mask of the local input. With $\mathcal{L}_{p}$, we can see from Figure~\ref{fig:LocalLoss}(h)(i) that the aforementioned problems are solved.


To summarize, our final loss function $\mathcal{L}$ is
\begin{equation}\label{eq7}
\begin{aligned}
&\mathcal{L}(\mathcal{F}(X,U_g,U_l;\theta),U_g, U_l, Y) \\
&={ \mathcal{L}}_{ g }+{ \mathcal{L}}_{ s }+{\mathcal{L}}_{p}.
\end{aligned}
\end{equation}




\subsection{Color theme recommendation system}
\begin{figure*}
  \centering
  \includegraphics[width=0.9\textwidth]{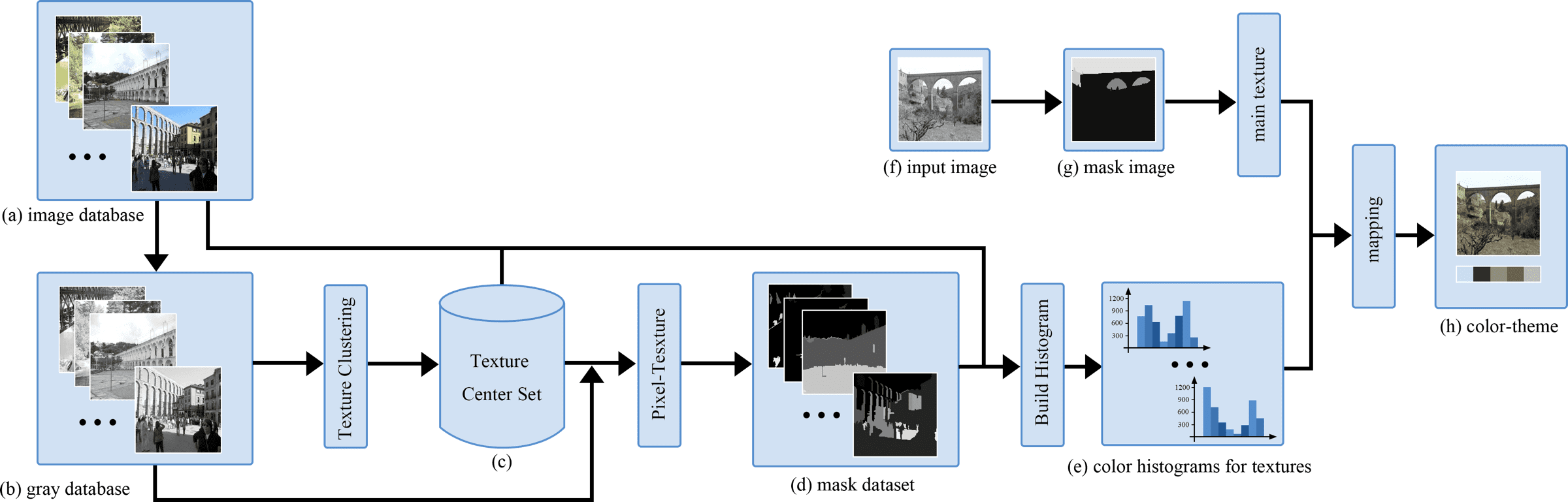}\\
  \caption{The overall pipeline of color theme commended system framework. The (c) is a texture library which stores the texture center of 120 textures. The image in (d) is a mapping between each pixel and texture, so the value of the pixel is the texture type of the pixel belongs. In the same way, (f) can be converted into (g) by calculating the nearest texture center (from (c)) of each segmentation. So the main texture in (f) can be counted. Combine with the main texture, the desired color theme into consideration to generate final recolored image in (h) can be found from (e).}\label{overall pipeline}
\end{figure*}

To help user determine the possible color theme of a gray-scale image, we propose a color theme recommending system.
Our system is inspired by the data driven method in \cite{Wang:2010:DIC:1882262.1866172}, which consists of an off-line process to builds a texture-color distribution mapping and an on-line process to lookup the color distribution for the segmented input color image. 
Different with the off-line process in \cite{Wang:2010:DIC:1882262.1866172}, which built its texture-color distribution mapping on color images to constrain the color distribution of textures, we build the mapping on the gray-color image pairs to predict the colors themes.





The framework of color theme commended system is illustrated in Figure~\ref{overall pipeline}. It consists of an off-line phase(lower row) and an online phase(upper row). The mapping of texture-color distribution is obtained in the off-line stage. The color themes of gray-scale image that input by users can be predicted by computing the main texture of gray-scale image in online phase.

During the off-line phase, we start with building an image database, by randomly choosing thousands of colorful images from Imagenet, and then convert the colorful images to grayscale images to form a grayscale image database. Next we adapt the graph-based method~\cite{Chia2011Semantic} to segment each gray-scale image into segmentations. We use the mean and standard deviation of the
pixelwise Gabor filter (scale=4, rotation=6) responses within every segmentation, resulting a 24 dimensional texture descriptor for every segmentation. We then build a texture library by clustering all the texture features into 120 textures with k-means algorithm. For each cluster, we accumulate a 2D color distribution with 100 bins by quantizing the corresponding {\em ab} channels of the segmentations. Hence, we build a texture-color distribution mapping.

\begin{figure*}
  \centering
  \begin{tabular} {@{\extracolsep{1mm}}c @{\extracolsep{0mm}}c @{\extracolsep{1mm}}c @{\extracolsep{0mm}}c @{\extracolsep{1mm}}c @{\extracolsep{0mm}}c @{\extracolsep{1mm}}c @{\extracolsep{0mm}}c @{\extracolsep{1mm}}c @{\extracolsep{0mm}}c }
		
		\includegraphics[width=0.095\textwidth]{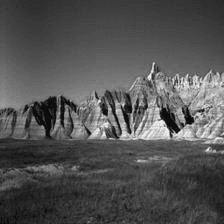} &
		\includegraphics[width=0.095\textwidth]{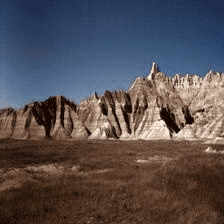} &
		\includegraphics[width=0.095\textwidth]{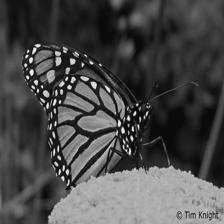} &
		\includegraphics[width=0.095\textwidth]{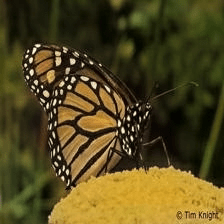} &
		\includegraphics[width=0.095\textwidth]{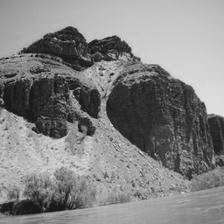} &
		\includegraphics[width=0.095\textwidth]{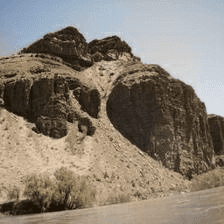} &
		\includegraphics[width=0.095\textwidth]{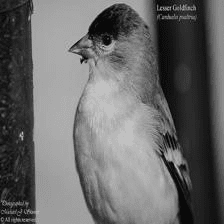} &
		\includegraphics[width=0.095\textwidth]{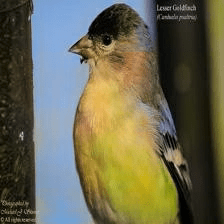} &
		\includegraphics[width=0.095\textwidth]{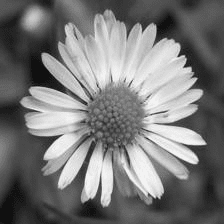} &
		\includegraphics[width=0.095\textwidth]{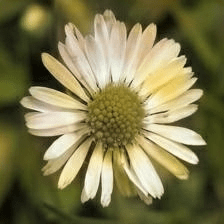} \\
		\includegraphics[width=0.095\textwidth]{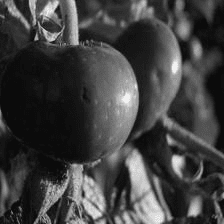} &
		\includegraphics[width=0.095\textwidth]{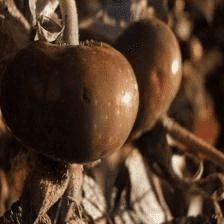} &
		\includegraphics[width=0.095\textwidth]{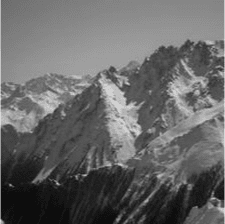} &
		\includegraphics[width=0.095\textwidth]{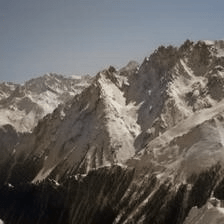} &
		\includegraphics[width=0.095\textwidth]{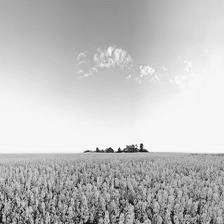} &
		\includegraphics[width=0.095\textwidth]{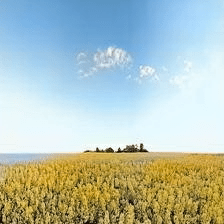} &
		\includegraphics[width=0.095\textwidth]{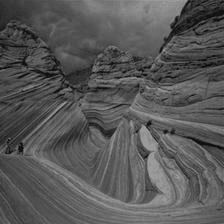} &
		\includegraphics[width=0.095\textwidth]{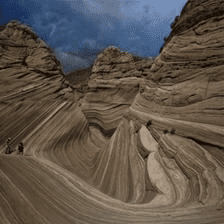} &
		\includegraphics[width=0.095\textwidth]{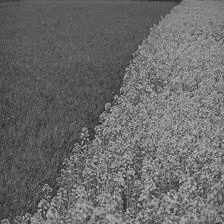} &
		\includegraphics[width=0.095\textwidth]{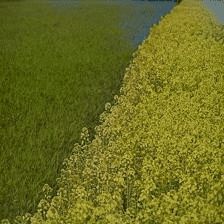} \\
		\includegraphics[width=0.095\textwidth]{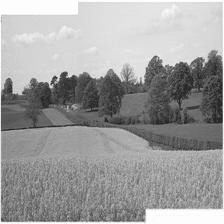} &
		\includegraphics[width=0.095\textwidth]{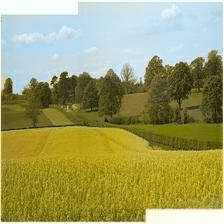} &
		\includegraphics[width=0.095\textwidth]{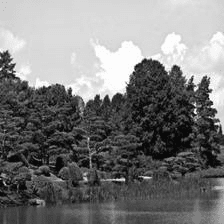} &
		\includegraphics[width=0.095\textwidth]{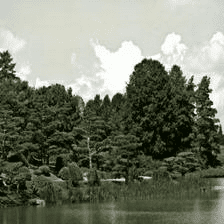} &
		\includegraphics[width=0.095\textwidth]{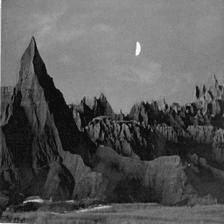} &
		\includegraphics[width=0.095\textwidth]{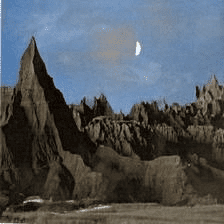} &
		\includegraphics[width=0.095\textwidth]{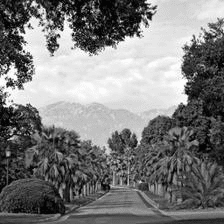} &
		\includegraphics[width=0.095\textwidth]{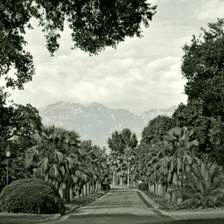} &
		\includegraphics[width=0.095\textwidth]{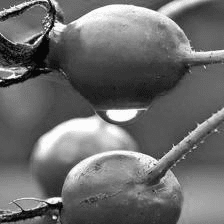} &
		\includegraphics[width=0.095\textwidth]{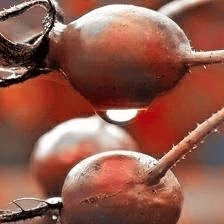} \\
		\includegraphics[width=0.095\textwidth]{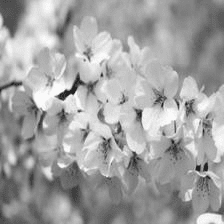} &
		\includegraphics[width=0.095\textwidth]{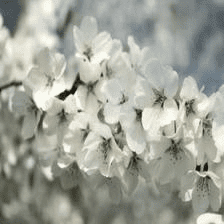} &
		\includegraphics[width=0.095\textwidth]{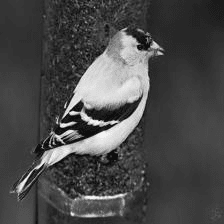} &
		\includegraphics[width=0.095\textwidth]{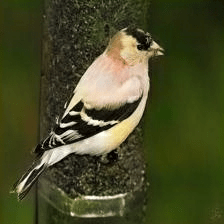} &
		\includegraphics[width=0.095\textwidth]{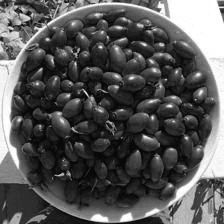} &
		\includegraphics[width=0.095\textwidth]{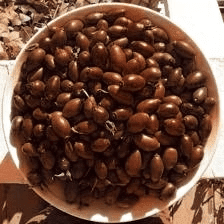} &
		\includegraphics[width=0.095\textwidth]{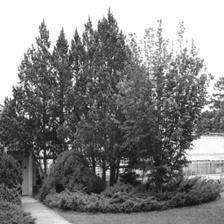} &
		\includegraphics[width=0.095\textwidth]{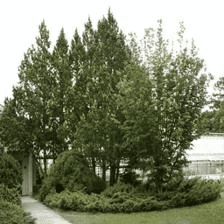} &
		\includegraphics[width=0.095\textwidth]{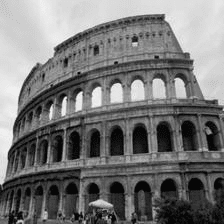} &
		\includegraphics[width=0.095\textwidth]{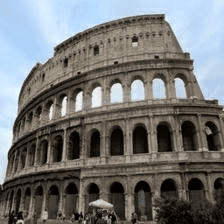} \\
		\includegraphics[width=0.095\textwidth]{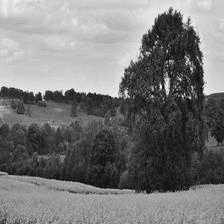} &
		\includegraphics[width=0.095\textwidth]{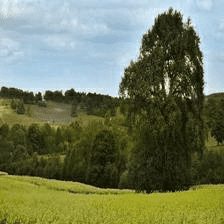} &
		\includegraphics[width=0.095\textwidth]{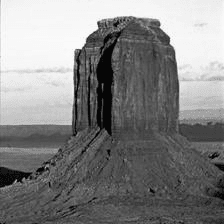} &
		\includegraphics[width=0.095\textwidth]{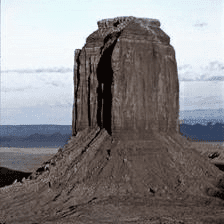} &
		\includegraphics[width=0.095\textwidth]{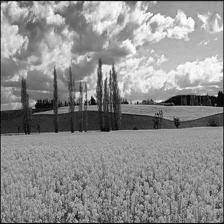} &
		\includegraphics[width=0.095\textwidth]{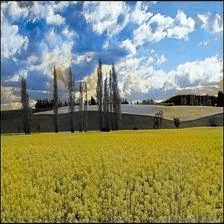} &
		\includegraphics[width=0.095\textwidth]{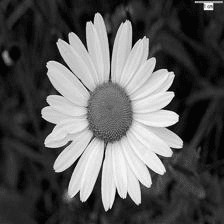} &
		\includegraphics[width=0.095\textwidth]{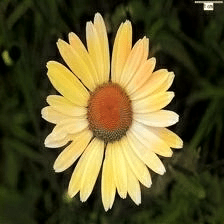} &
		\includegraphics[width=0.095\textwidth]{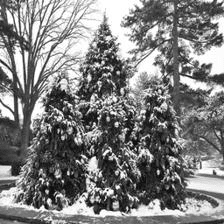} &
		\includegraphics[width=0.095\textwidth]{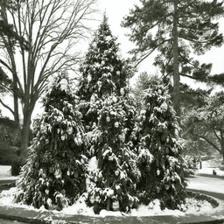} \\
		\includegraphics[width=0.095\textwidth]{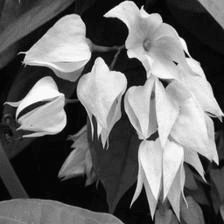} &
		\includegraphics[width=0.095\textwidth]{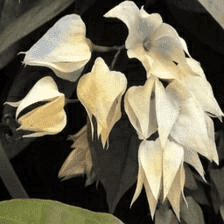} &
		\includegraphics[width=0.095\textwidth]{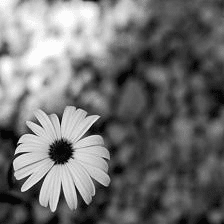} &
		\includegraphics[width=0.095\textwidth]{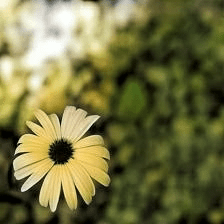} &
		\includegraphics[width=0.095\textwidth]{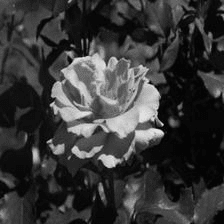} &
		\includegraphics[width=0.095\textwidth]{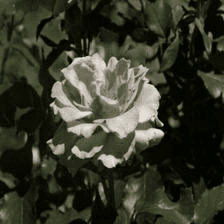} &
		\includegraphics[width=0.095\textwidth]{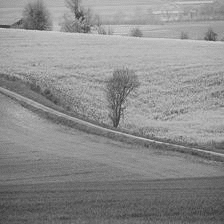} &
		\includegraphics[width=0.095\textwidth]{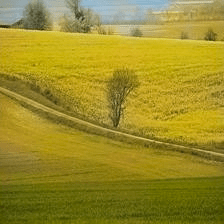} &
		\includegraphics[width=0.095\textwidth]{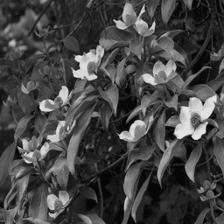} &
		\includegraphics[width=0.095\textwidth]{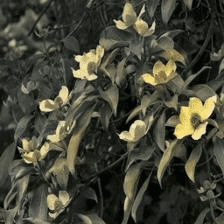} \\
		
	\end{tabular}
  \caption{Examples of recommendation system. }\label{RecommendationExamples}
\end{figure*}

In the online-phase, we segment the input gray-scale image, extract the texture feature vectors, find the closet clusters in the texture library, and lookup the corresponding color distributions. We choose the most frequent color of the K largest segmentations (with more number of pixels) as our recommended color theme.
 We display some images colored by color themes which are generated by this system in Figure~\ref{RecommendationExamples}.

\section{Experiment}



\begin{figure}
  \centering
  \begin{tabular} {@{\extracolsep{1mm}}c @{\extracolsep{1mm}}c @{\extracolsep{1mm}}c @{\extracolsep{2mm}}c @{\extracolsep{1mm}}c @{\extracolsep{1mm}}c }
		
		\includegraphics[width=0.07\textwidth]{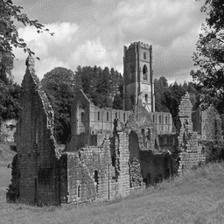} &
		\includegraphics[width=0.07\textwidth]{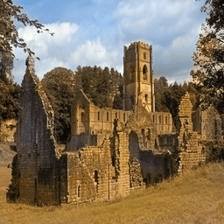} &
		\includegraphics[width=0.07\textwidth]{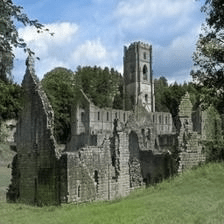} &
		\includegraphics[width=0.07\textwidth]{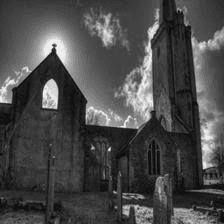} &
		\includegraphics[width=0.07\textwidth]{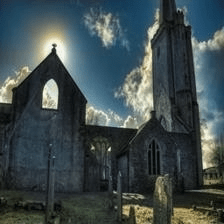} &
		\includegraphics[width=0.07\textwidth]{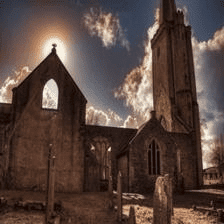} \\
		
		{} &
		\includegraphics[width=0.07\textwidth]{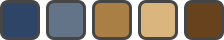} &
		\includegraphics[width=0.07\textwidth]{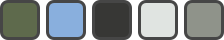} &
		{} &
		\includegraphics[width=0.07\textwidth]{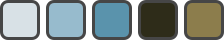} &
		\includegraphics[width=0.07\textwidth]{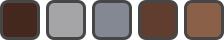} \\
		\\
		
		\includegraphics[width=0.07\textwidth]{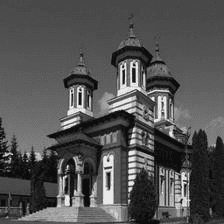} &
		\includegraphics[width=0.07\textwidth]{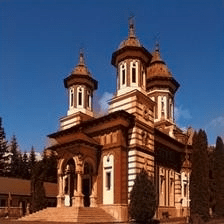} &
		\includegraphics[width=0.07\textwidth]{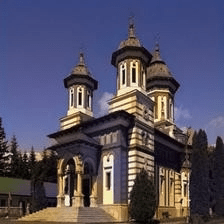} &
		\includegraphics[width=0.07\textwidth]{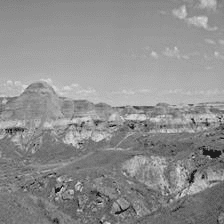} &
		\includegraphics[width=0.07\textwidth]{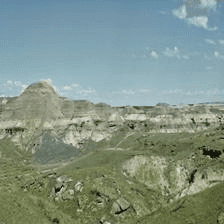} &
		\includegraphics[width=0.07\textwidth]{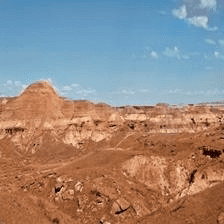} \\
		
		{} &
		\includegraphics[width=0.07\textwidth]{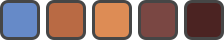} &
		\includegraphics[width=0.07\textwidth]{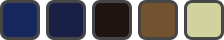} &
		{} &
		\includegraphics[width=0.07\textwidth]{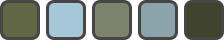} &
		\includegraphics[width=0.07\textwidth]{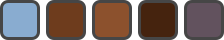} \\

	\end{tabular}
  \caption{Examples of colorization results in the training set Place.}\label{Place result}
\end{figure}

\begin{figure*}
\centering
	\begin{tabular}{@{\extracolsep{1mm}}c @{\extracolsep{1mm}}c @{\extracolsep{1mm}}c @{\extracolsep{1mm}}c @{\extracolsep{1mm}}c @{\extracolsep{1mm}}c @{\extracolsep{1mm}}c @{\extracolsep{1mm}}c}
	
	        {Gray Image} & ~\cite{Efros2017Real} & ~\cite{IizukaSIGGRAPH2016} & ~\cite{Zhang2016Colorful} & {Ours} &  {Ground Truth} \\
	
	         \\ \noalign {\vspace{0.5cm}}
		
		 {} & {16.5179} & \textbf{21.6247} & {18.28} & {16.7994} & {} \\
		\includegraphics[width=0.16\textwidth]{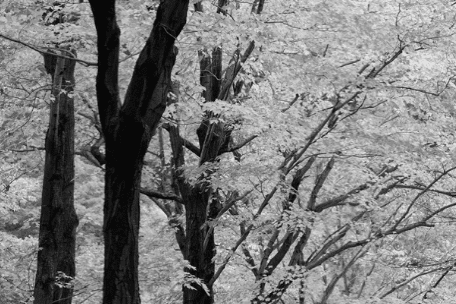} &
		\includegraphics[width=0.16\textwidth]{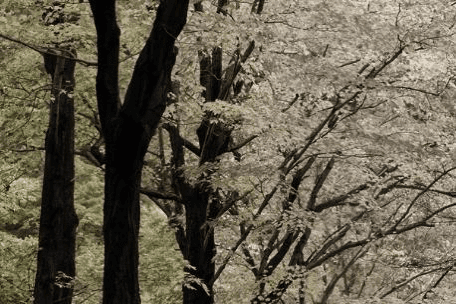} &
		\includegraphics[width=0.16\textwidth]{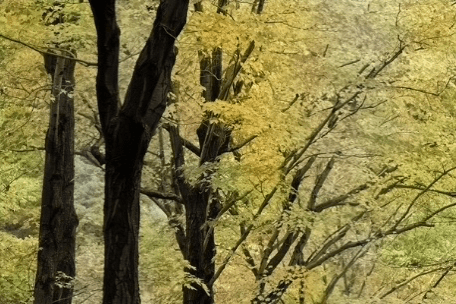} &
		\includegraphics[width=0.16\textwidth]{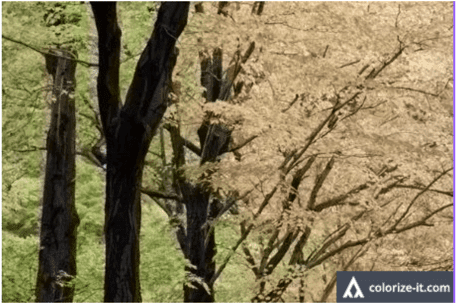} &
		\includegraphics[width=0.16\textwidth]{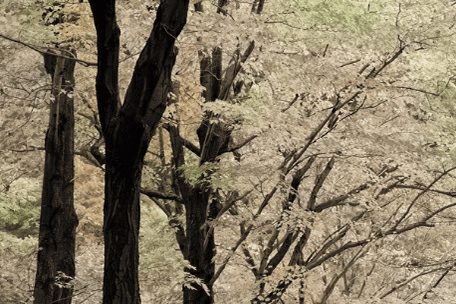} &
		\includegraphics[width=0.16\textwidth]{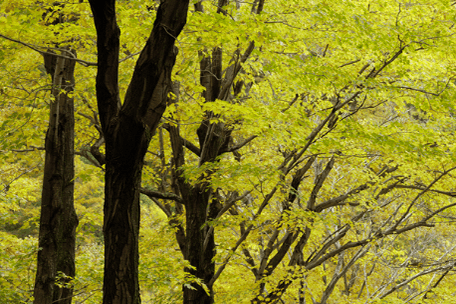} \\
		\\ \noalign {\vspace{0.3mm}}
		
		 {} & {28.4547} & {23.7707} & {22.6357} & \textbf{29.0915} & {} \\
		\includegraphics[width=0.16\textwidth]{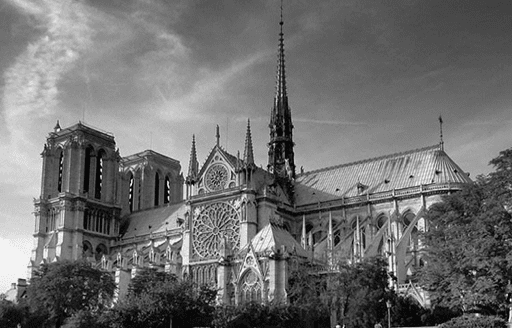} &
		\includegraphics[width=0.16\textwidth]{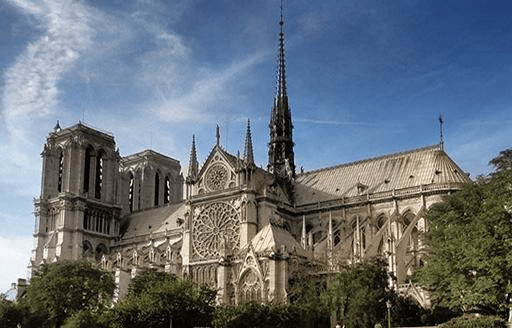} &
		\includegraphics[width=0.16\textwidth]{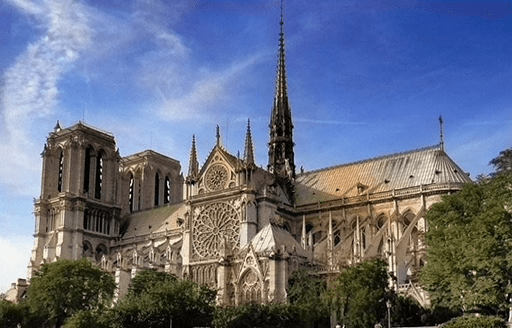} &
		\includegraphics[width=0.16\textwidth]{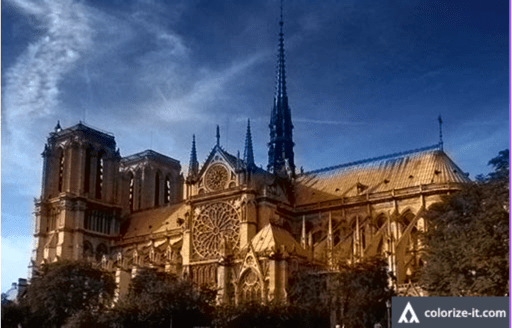} &
		\includegraphics[width=0.16\textwidth]{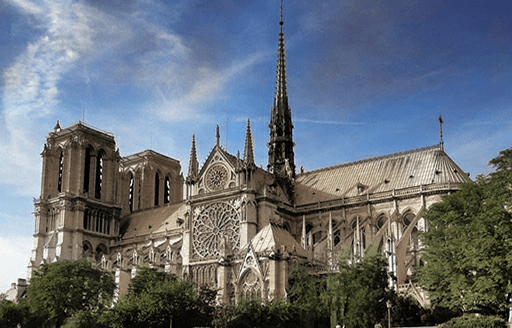} &
		\includegraphics[width=0.16\textwidth]{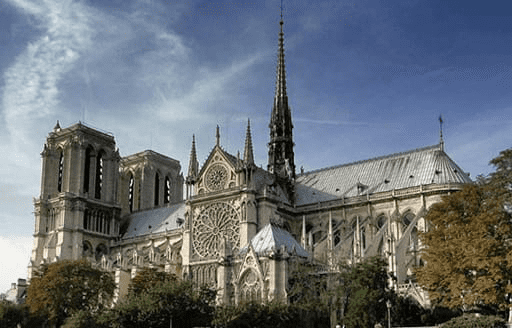} \\
		\\ \noalign {\vspace{0.3mm}}
		
		 {} & {19.4188} & \textbf{21.2468} & {20.3247} & {19.8622} & {} \\
		\includegraphics[width=0.16\textwidth]{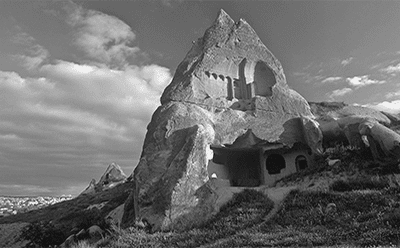} &
		\includegraphics[width=0.16\textwidth]{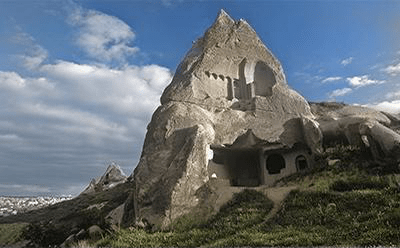} &
		\includegraphics[width=0.16\textwidth]{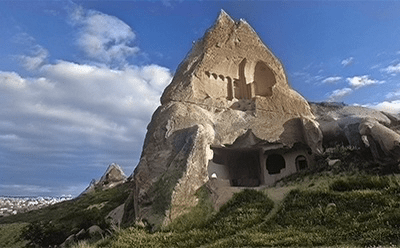} &
		\includegraphics[width=0.16\textwidth]{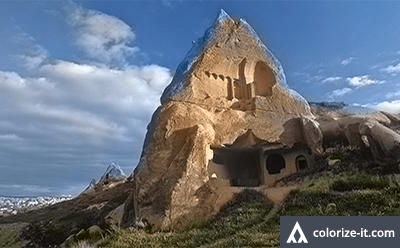} &
		\includegraphics[width=0.16\textwidth]{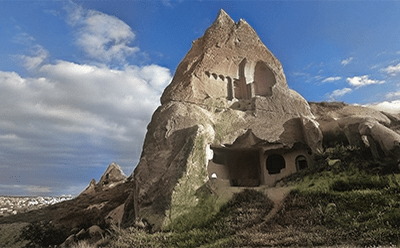} &
		\includegraphics[width=0.16\textwidth]{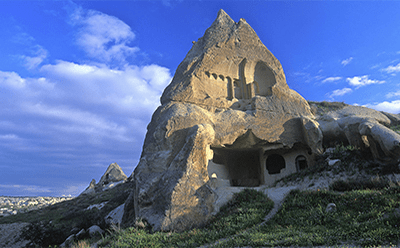} \\
		\\ \noalign {\vspace{0.3mm}}
		
		 {} & {13.931} & {14.9813} & {10.6654} & \textbf{16.6442} & {} \\
		\includegraphics[width=0.16\textwidth]{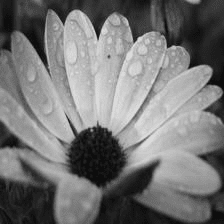} &
		\includegraphics[width=0.16\textwidth]{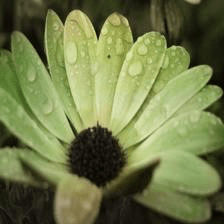} &
		\includegraphics[width=0.16\textwidth]{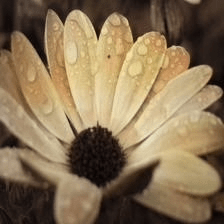} &
		\includegraphics[width=0.16\textwidth]{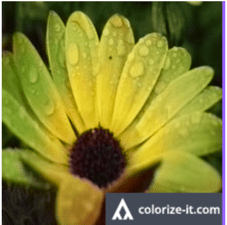} &
		\includegraphics[width=0.16\textwidth]{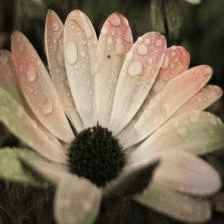} &
		\includegraphics[width=0.16\textwidth]{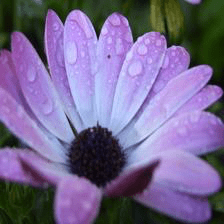} \\
		\\ \noalign {\vspace{0.3mm}}
		
		{} & {25.0427} & {23.4359} & {19.9974} & \textbf{26.0092} & {} \\
		\includegraphics[width=0.16\textwidth]{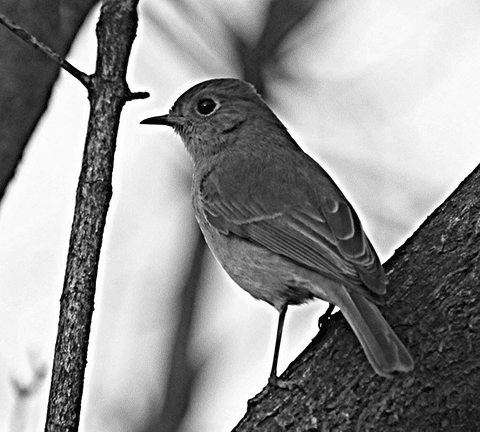} &
		\includegraphics[width=0.16\textwidth]{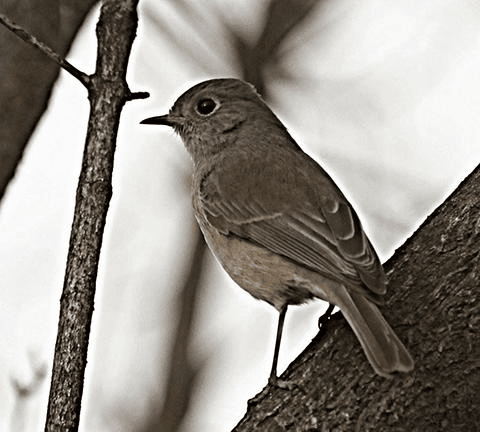} &
		\includegraphics[width=0.16\textwidth]{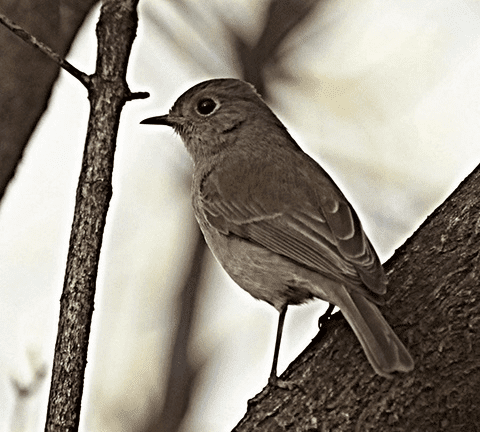} &
		\includegraphics[width=0.16\textwidth]{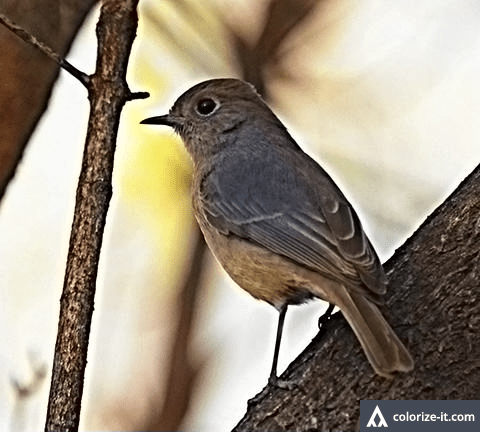} &
		\includegraphics[width=0.16\textwidth]{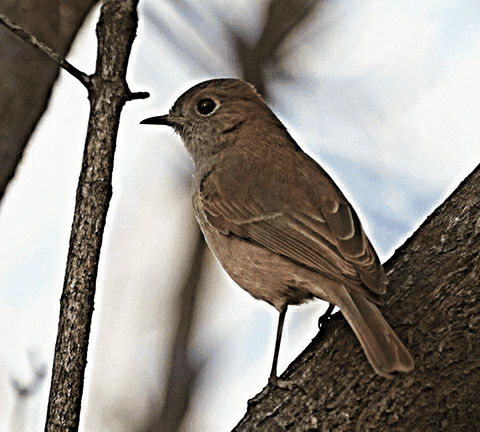} &
		\includegraphics[width=0.16\textwidth]{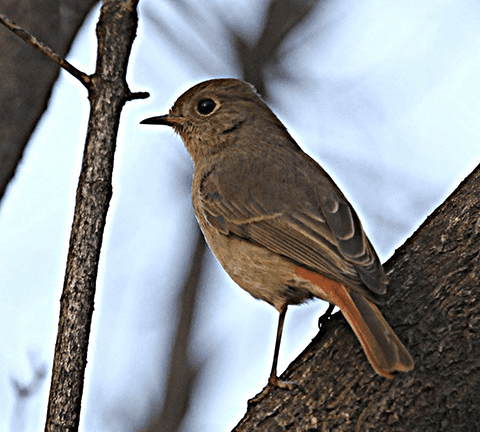} \\
		
		(a) & (b) & (c) & (d) & (e) & (f) \\

	\end{tabular}	
	\caption{Comparison with automatic colorization methods. (a) are grayscale images. (b) to (e) are results of automatic colorization methods. (f) are ground truth images. The digits above every image are PSNR, in which the bold one is the best.}\label{Comparison with previous method}
\end{figure*}

\begin{figure*}
  \centering
\begin{tabular}{@{\extracolsep{0mm}}c @{\extracolsep{1mm}}c @{\extracolsep{1mm}}c @{\extracolsep{1mm}}c @{\extracolsep{1mm}}c @{\extracolsep{1mm}}c @{\extracolsep{1mm}}c @{\extracolsep{1mm}}c @{\extracolsep{1mm}}c}
	
	         Input &
		\includegraphics[height=1.75cm]{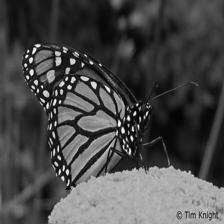} &
		{} &
		{} &
		\includegraphics[height=1.75cm]{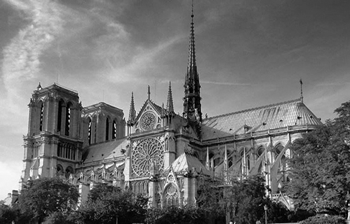} &
		\includegraphics[height=1.75cm]{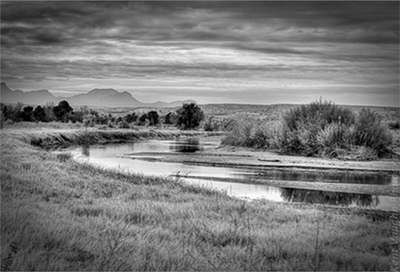} &
		\includegraphics[height=1.75cm]{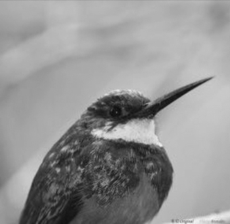} &
		\includegraphics[height=1.75cm]{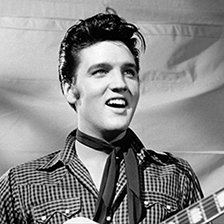} &
		\includegraphics[height=1.75cm]{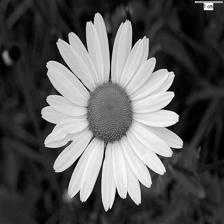} \\
		
		{} &
		{} &
		{} &
		{} &
		\includegraphics[height=3mm]{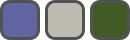} &
		\includegraphics[height=3mm]{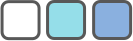} &
		\includegraphics[height=3mm]{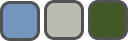} &
		\includegraphics[height=3mm]{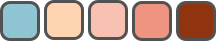} &
		\includegraphics[height=3mm]{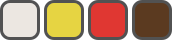} \\
		
		Local &
		\includegraphics[height=1.75cm]{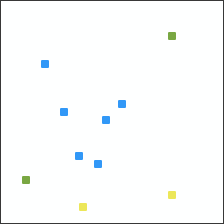} &
		\includegraphics[height=1.75cm]{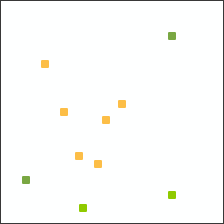} &
		\includegraphics[height=1.75cm]{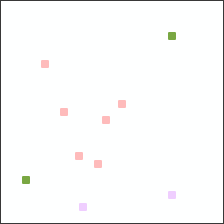} &
		\includegraphics[height=1.75cm]{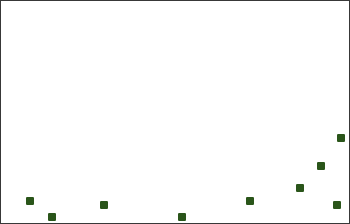} &
		\includegraphics[height=1.75cm]{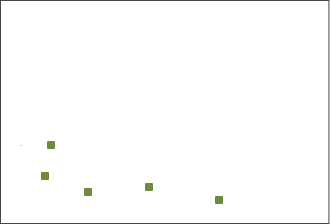} &
		\includegraphics[height=1.75cm]{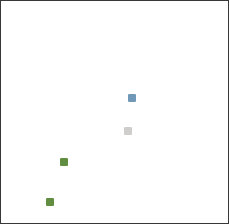} &
		\includegraphics[height=1.75cm]{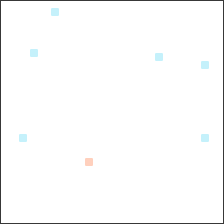} &
		\includegraphics[height=1.75cm]{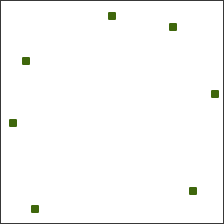} \\
		
		Ours &
		\includegraphics[height=1.75cm]{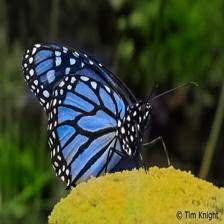} &
		\includegraphics[height=1.75cm]{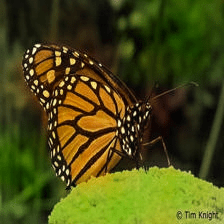} &
		\includegraphics[height=1.75cm]{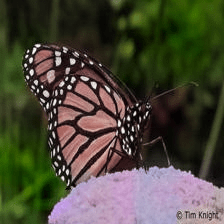} &
		\includegraphics[height=1.75cm]{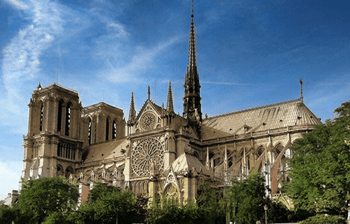} &
		\includegraphics[height=1.75cm]{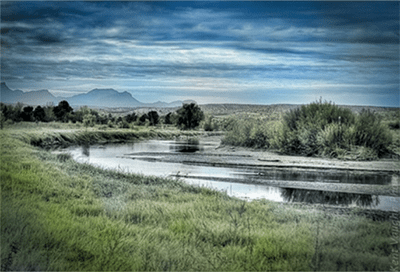} &
		\includegraphics[height=1.75cm]{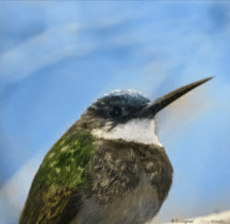} &
		\includegraphics[height=1.75cm]{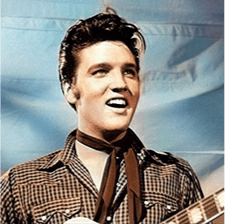} &
		\includegraphics[height=1.75cm]{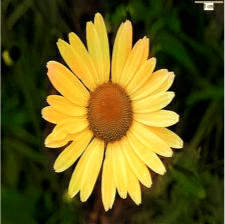} \\
		
		~\cite{Efros2017Real} &
		\includegraphics[height=1.75cm]{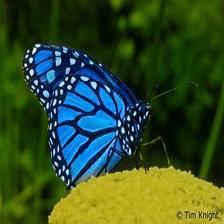} &
		\includegraphics[height=1.75cm]{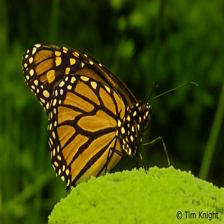} &
		\includegraphics[height=1.75cm]{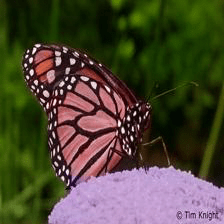} &
		\includegraphics[height=1.75cm]{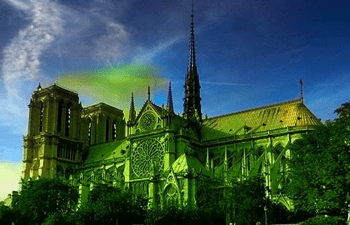} &
		\includegraphics[height=1.75cm]{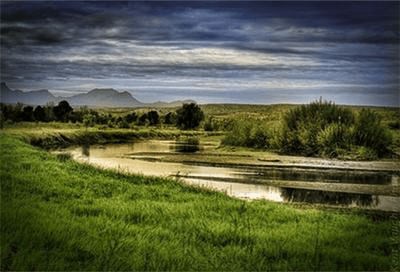} &
		\includegraphics[height=1.75cm]{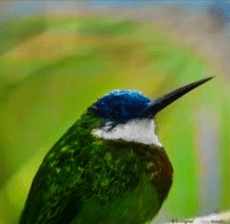} &
		\includegraphics[height=1.75cm]{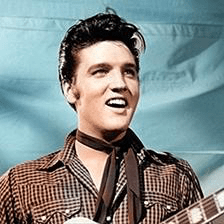} &
		\includegraphics[height=1.75cm]{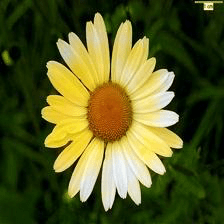} \\
		
		~\cite{Levin2004Colorization} &
		\includegraphics[height=1.75cm]{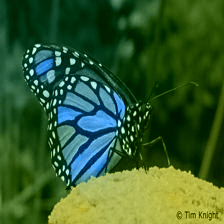} &
		\includegraphics[height=1.75cm]{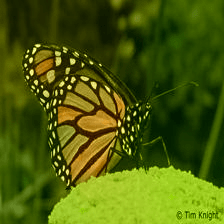} &
		\includegraphics[height=1.75cm]{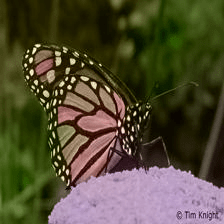} &
		\includegraphics[height=1.75cm]{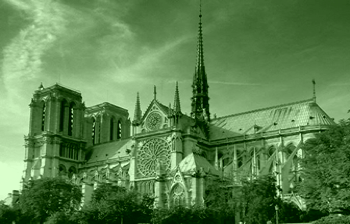} &
		\includegraphics[height=1.75cm]{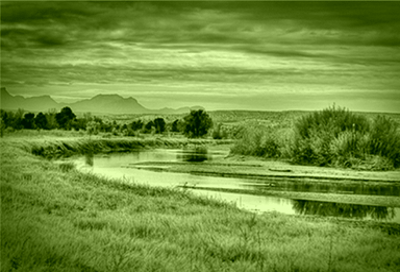} &
		\includegraphics[height=1.75cm]{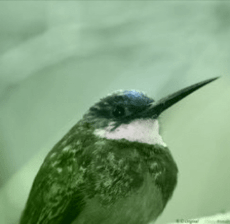} &
		\includegraphics[height=1.75cm]{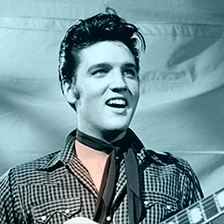} &
		\includegraphics[height=1.75cm]{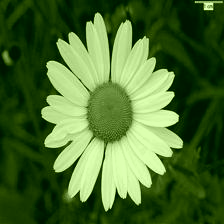} \\
				
	\end{tabular}
  \caption{Comparisons of interactive colorization methods. We only add local inputs on the first grayscale image and change the color of some points on the Butterfly wings, without changing their position. For the last five images, we add color theme and local inputs simultaneously, while other methods add local inputs only.}\label{Comparison with methods}
\end{figure*}

\begin{figure}[h]
\centering
	\begin{tabular} {@{\extracolsep{1mm}}c @{\extracolsep{1mm}}c }
	
		\includegraphics[height=0.5cm]{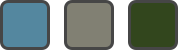} &
		{} \\
		
		\includegraphics[width=0.23\textwidth]{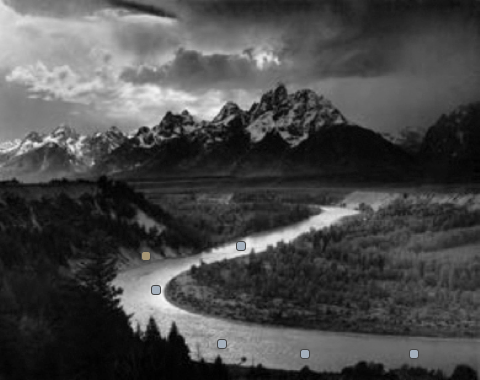} &
		\includegraphics[width=0.23\textwidth]{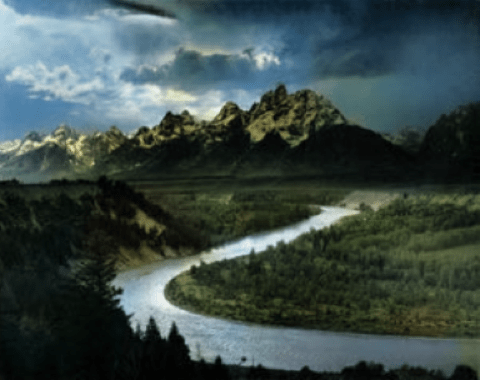} \\
		
		\includegraphics[height=0.5cm]{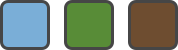} &
		{} \\
		
		\includegraphics[width=0.23\textwidth]{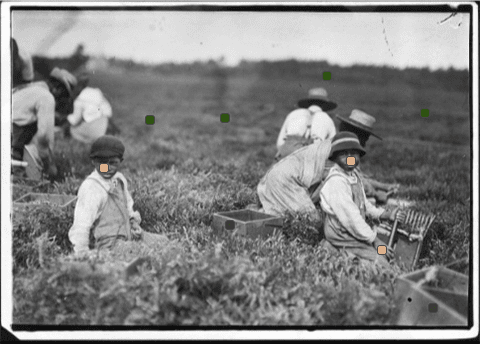} &
		\includegraphics[width=0.23\textwidth]{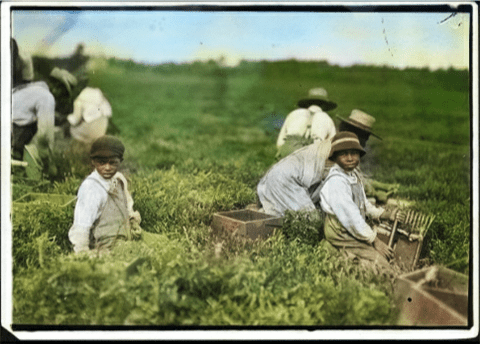} \\
		
		\includegraphics[width=0.23\textwidth]{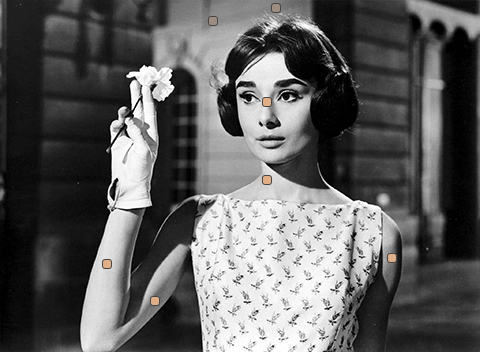} &
		\includegraphics[width=0.23\textwidth]{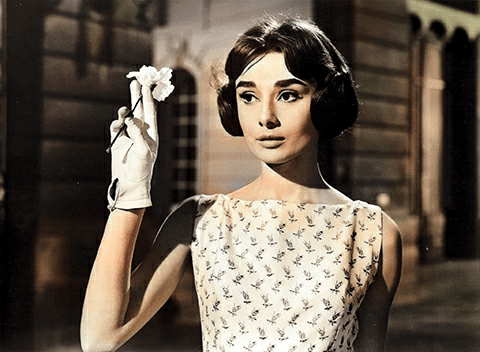} \\

	\end{tabular}
	\caption{Examples for coloring historic images. The first two images are colored by color themes and local inputs. The last image is colored by local inputs only.}\label{ColorForPast}
\end{figure}

We implemented the network model on the NIVIDIA GTX1080Ti GPU, using Google's TensorFlow architecture. We use the Place dataset to train our model. Figure~\ref{Place result} shows some colored results of this dataset. The K-color map images, color themes, local inputs and their corresponding masks are generated on Matlab. Then, we randomly adjust the order of training sets as input of our model. We train the model using a batch size of 50 for 40,000 iterations, which takes about two days on the GPU. To enable the four combinations of input: no inputs, only global input, only local input, and both global and local input, we randomly select one kind of combination in the training. The ratio to select each combination is equal to $25\%$.


In the previous works, the normalization of input has been shown to speed up the learning speed. In {\em Lab} space, the range of {\em L} is $[0-100]$, and the range of {\em ab} channel is $[127, -128]$. In order to accelerate the learning speed, we normalize the range of {\em L} and {\em ab} to [0,1] as
\begin{equation}\label{eq8}
\begin{array}{l}
 L = L/100 \\
 ab = ( ab + 128 ) / 255
\end{array}
\end{equation}

\subsection{Automatic Colorization}\label{GrayscaleImageColorization}
In these section, we compare our automatic colorization method with several state-of-art methods, including ~\cite{Efros2017Real}, which train coloring with no inputs or local inputs , ~\cite{IizukaSIGGRAPH2016} and ~\cite{Zhang2016Colorful}, which focus on automatic colorization. The colorized results of these four methods are shown in Figure~\ref{Comparison with previous method}. As shown in the figure, all the automatic methods can colorize the gray-scale images. But due to the different training sets, the colorized images may have different color styles. In terms of numerical evaluation, our method and ~\cite{IizukaSIGGRAPH2016} get better PSNRs without compared with the other two methods. All PSNRs of our automatic results are higher than ~\cite{Efros2017Real}. Note that, although automatic colorization is only our by-product, it still has good performances.

\subsection{Interactive Colorization}\label{ColorizationWithTwoInput}

In this section, we compare our interactive colorization method with the state-of-art user guide method~\cite{Efros2017Real}, and the classical optimization method~\cite{Levin2004Colorization}. The results are shown in Figure~\ref{Comparison with methods}. The Butterfly images in the first three columns show results colorized by different local input points. With only local inputs, the results of our method are similar to those of the user-guide method~\cite{Efros2017Real}. In contrast, the results of \cite{Levin2004Colorization} fail to diffuse the color to the whole butterfly.


For the next five images, our method uses global input (color theme) and local input together.  The other two methods use local inputs only. For images with less clear boundaries, like the fourth column, our method can faithfully assign green to the trees, where the other methods cause color overflows to the background.
Using our method, users can control the colorized image by choosing a color theme, and add a small number of local input points to certain regions, while other methods maybe need to add more local input points to produce the same results. For example, in the fourth column, users maybe need to add several extra color points on the house and sky to avoid color overflow and assign certain colors for the user guide method~\cite{Efros2017Real}. ~\cite{Levin2004Colorization} needs to assign more color points for everywhere in every image. Otherwise, the whole image is going to be one color as shown in the figure. Besides, our method also shows good performance on other type images, like human and natural pictures showed in the last three columns of the Figure~\ref{Comparison with methods}.

In general, our method provides a more convenient and efficient tool for the users by supporting four combinations of input in a single network: no inputs, only global inputs, only local inputs, and simultaneous global input and local input.


\subsection{Numerical Comparisons}

Besides visual comparisons, we also evaluate the PSNRs of the related methods. We randomly choose another 20 images and colorize them using the automatic methods and interactive methods mentioned above. We use original images as global inputs for ~\cite{Efros2017Real}, and use color themes extracted from original images as global inputs for our method. Local inputs are 3 to 20 colorful points randomly extracted from original images. Then we calculate the average PSNR of them as showed in Table~\ref{tab1}. Through Table~\ref{tab1}, we can see that results of automatic methods have similar performances, which have almost identical PSNRs. When we add global inputs, our method works much better and gets a higher PSNR.


For method with global inputs, the user guide method~\cite{Efros2017Real} shows good performance and get the highest PSNR. Our method also works well. But the PSNR of method is a bit lower than the user guide method~\cite{Efros2017Real}. This is possibly because we train our network to support four combination of inputs , the effect of local inputs is not optimized in our method.

For method with local inputs, the user guide method~\cite{Efros2017Real} shows good performance and get the highest PSNR.  Our method also works well. But the PSNR of method is a bit lower than the user guide method~\cite{Efros2017Real}. This is possibly also because we train our network to support four combination of inputs , the effect of local inputs is not optimized in our method. The optimization method ~\cite{Levin2004Colorization} gets a lower PSNR, probably because it was designed for strokes not point inputs.
  When both global inputs and local inputs  are used, our method get a higher PSNR, which is very close to the highest in Table~\ref{tab1}.


\begin{table}[t]
\begin{center}
\caption{The average PSNR of 20 images.} \label{tab1}
\centering
\begin{tabular}{|p{2cm}<{\centering}|p{3cm}<{\centering}|p{2cm}<{\centering}|}
  \hline
 \textbf{Method} & \textbf{Added Inputs} & \textbf{PSNR(dB)} \\
  \hline
  ~\cite{Zhang2016Colorful} & automatic  &  23.8907 \\
  ~\cite{IizukaSIGGRAPH2016} & automatic  &  24.4463 \\
  ~\cite{Efros2017Real} & automatic  &  24.4997 \\
  Ours & automatic  &  24.4691 \\
  \hline
  ~\cite{Efros2017Real} &  global inputs &  29.2105 \\
  Ours &  global inputs  &  27.8543 \\
  \hline
  ~\cite{Levin2004Colorization} &  local inputs  &  24.075 \\
  ~\cite{Efros2017Real} &  local inputs  &  29.0978 \\
  Ours &  local inputs  &  27.2619 \\
  \hline
  Ours &  global + local  &  28.5375 \\
  \hline
\end{tabular}
\end{center}
\end{table}

\subsection{Colorization for past}
We test our method on some historic images, which are white-and-black images. Although these images are different from our dataset, as that they may have rough edges and their picture quality may be unclear, they also can be colorized by our method. Compared with our method, other methods, which use local input only, maybe need a lot of local input points to get the same results. Some examples of coloring historic images are shown in Figure~\ref{ColorForPast}.

\subsection{Computation time}
We test computation time of two image sizes on the NIVIDIA GTX1080Ti GPU. We take the average time of 100 computations on 50 images to get a reliable testing value, as shown in Table~\ref{ComputationTime}. The data shows that our method is comparable to real-time colorization.

\begin{table}[t]
\begin{center}
\caption{The average computation time.} \label{ComputationTime}
\begin{tabular}{|c|c|c|c|}
  \hline
  Image Size & Pixels & Time \\
  \hline
  $256\times 256$ & 65,536  &  \textbf{0.00750 s} \\
  \hline
   $512\times 512$ & 262,144  &  \textbf{0.0228 s} \\
  \hline
\end{tabular}
\end{center}
\end{table}

\subsection{Limitations and discussion}
Our model is trained by the Place dataset, which can not include everything in this world. If an gray-scale image contains certain contents, which is not learned from the data, the colors may be not assigned appropriately. In this case, the user may need to give more inputs to the system. For some images with unclear edges, the network may produce unexpected results, like color overflow on background. Users can add another local input points in the background region. In addition, When users add local input points with unusual colors, they may have to add more points to achieve the desired effect or directly change to use an appropriate color theme.


\section{Conclusion and Future Work}
In this paper, we propose a novel interactive deep colorization method. By designing a suitable loss function, our method allows four combinations of input in a single network model, including no inputs, only global input, only local input, and both global input and local input. Furthermore, our global input is a color theme with variable number of colors, which is more easy and straightforward for user to use. Together with recommendation system, our method can reduce time and cost of coloring images for users. In addition, we evaluate our method on images of outdoor, human picture, past pictures, and show that it can produce satisfactory results for all them.
In the future, we would like to extend our method to gray-scale video sequence.




\bibliographystyle{IEEEbib}

\end{document}